\documentclass[11pt]{article}

\usepackage[preprint]{acl}

\usepackage{times}
\usepackage{latexsym}
\usepackage[T1]{fontenc}
\usepackage[utf8]{inputenc}
\usepackage{microtype}
\usepackage{inconsolata}
\usepackage{graphicx}

\usepackage{amsmath}
\usepackage{amssymb}
\usepackage{booktabs}
\usepackage{tabularx}
\usepackage{array}
\usepackage{multirow}
\usepackage{enumitem}
\usepackage{xcolor}
\usepackage{listings}
\usepackage{upquote}
\usepackage[skins,breakable,listings]{tcolorbox}
\usepackage{tikz}
\usetikzlibrary{arrows.meta}

\definecolor{cardBg}{HTML}{F8FAFC}
\definecolor{cardHeaderBg}{HTML}{F1F5F9}
\definecolor{cardBorder}{HTML}{CBD5E1}
\definecolor{inkPrimary}{HTML}{0F172A}
\definecolor{inkMuted}{HTML}{64748B}
\definecolor{codeKw}{HTML}{1E3A8A}
\definecolor{codeStr}{HTML}{0F766E}
\definecolor{codeCom}{HTML}{64748B}
\definecolor{warnCrimson}{HTML}{991B1B}
\definecolor{gainForest}{HTML}{0F766E}

\lstdefinestyle{minimalistBase}{
  basicstyle=\footnotesize\ttfamily\linespread{1.04}\selectfont,
  columns=fullflexible,
  keepspaces=true,
  upquote=true,
  showstringspaces=false,
  breaklines=true,
  breakatwhitespace=true,
  postbreak=\mbox{\textcolor{inkMuted}{$\hookrightarrow$}\space},
  aboveskip=0pt,
  belowskip=0pt,
  literate=
    {—}{{---}}1
    {–}{{--}}1
    {…}{{...}}1
    {→}{{->}}1
    {×}{{x}}1
    {−}{{-}}1
    {≥}{{>=}}1
    {≤}{{<=}}1
}

\lstdefinestyle{minimalistPython}{
  style=minimalistBase,
  basicstyle=\scriptsize\ttfamily\linespread{1.02}\selectfont,
  language=Python,
  keywordstyle=\bfseries\color{codeKw},
  stringstyle=\color{codeStr},
  commentstyle=\color{codeCom},
  numbers=left,
  numberstyle=\tiny\ttfamily\color{inkMuted},
  numbersep=7pt,
  xleftmargin=12pt
}

\tcbset{
  minimalCard/.style={
    enhanced,
    breakable,
    pad at break*=4pt,
    colback=cardBg,
    colframe=cardBorder,
    colbacktitle=cardHeaderBg,
    coltitle=inkPrimary,
    boxrule=0.5pt,
    titlerule=0.5pt,
    arc=1.5pt,
    left=8pt, right=8pt, top=4pt, bottom=4pt,
    toptitle=3pt, bottomtitle=3pt,
    fonttitle=\footnotesize\ttfamily\bfseries,
    before skip=4.5pt,
    after skip=5.5pt
  }
}

\newtcolorbox{promptcard}[2]{
  minimalCard,
  title={#1\hfill{\normalfont\scriptsize\sffamily\color{inkMuted}#2}}
}

\newtcblisting{skilllisting}[2]{
  minimalCard,
  listing only,
  listing options={style=minimalistBase},
  title={#1\hfill{\normalfont\scriptsize\sffamily\color{inkMuted}#2}}
}

\newtcblisting{pylisting}[2]{
  minimalCard,
  listing only,
  listing options={style=minimalistPython},
  title={#1\hfill{\normalfont\scriptsize\sffamily\color{inkMuted}#2}}
}

\newcommand{\method}{\textsc{EfficientSkill}}
\newcommand{\evoskill}{\textsc{EvoSkill}}

\title{\method{}: Learning Skills for Token-Efficient Multi-Turn Agents}

\author{Deep Shah\textsuperscript{1}, Sanket Badhe\textsuperscript{1}, Nehal Kathrotia\textsuperscript{1}, Priyanka Tiwari\textsuperscript{2} \\[0.5em]
  \textsuperscript{1}Google LLC, \textsuperscript{2}Purdue University}

\begin{document}
\maketitle

\begin{abstract}
Reflective skill optimizers improve multi-turn LLM agents by revising instructions from failed and successful rollouts, yet accumulating edge-case rules bloats prompts, inflates token use, and pushes agents into long, error-prone loops. We introduce \method{}, a two-step framework that \emph{first} discovers a reliable skill without token penalties, and \emph{then} warm-starts accuracy-guarded pruning from that skill to compress instructions, bound agent turns, and offload reusable code into workspace utilities with zero implementation prompt cost. Across four frontier, mid-tier, and open-weight models on \textsc{MATH-500}, \textsc{ALFWorld}, and \textsc{SpreadsheetBench}, \method{} cuts output tokens over unconstrained \evoskill{} on $11$ of $12$ configurations (by up to $80\%$) while raising mean accuracy on every benchmark, including $+14.3$ and $+17.5$ percentage-point gains on \textsc{SpreadsheetBench} with Gemini 3.8 Flash and Grok 4.1 Fast by replacing stalled loops with compact workflows. By shrinking both generated tokens and replayed history, \method{} lowers total token usage and serving cost across every evaluated configuration, with or without full-prefix KV caching.
\end{abstract}

\section{Introduction}
\label{sec:intro}

Coding and tool-use agents increasingly organize domain knowledge into modular skill packages that pair natural-language instruction documents with executable workspace scripts and utilities \citep{li2026skillsbench,evoskill2026}. Recent reflective optimizers automate skill authoring by iteratively revising instructions from failed execution traces \citep{yuksekgonul2024textgrad,gepa2025,evoskill2026,moll2026grasp}, yet optimize almost exclusively for task accuracy.

In multi-turn deployments, optimizing skills for accuracy alone creates two compounding problems:
\begin{itemize}[leftmargin=1.2em, itemsep=1.2pt, topsep=1.5pt]
    \item \textbf{High Output Cost and Quadratic History Growth:} Output and reasoning tokens cost up to $5\times$ input rates and drive up per-turn billing. Moreover, because multi-turn APIs resend the growing interaction history at every turn, cumulative input tokens grow quadratically \citep{ali2026harness,liu2026contextcost}.
    \item \textbf{Prompt Over-Fitting and Multi-Turn Drift:} Repeatedly patching instructions for edge cases yields over-fitted prompts. Rather than acting decisively, agents fall into hesitant re-inspection loops, unproductive deliberation turns, and cascading execution errors \citep{laban2025lost,khanal2026beyond}.
\end{itemize}

Efficiency methods for prompts \citep{nayab2024concise,xu2025cod,han2025tale} and single-stage cost-regularized optimizers \citep{paretoprompt2025,shah2026crop} penalize token usage from the start of search. When an agent begins with a low-accuracy prompt on an unfamiliar environment, solving the task first requires exploratory instructions and diagnostic steps; penalizing tokens too early suppresses and stalls optimization near its starting accuracy. Conversely, post-hoc compressors such as SkillReducer \citep{gao2026skillreducer} trim static skill instructions ($-27\%$ input tokens) without shortening the reasoning or actions an agent generates at runtime.

\begin{figure*}[t]
    \centering
    \includegraphics[width=0.81\textwidth, trim=44bp 18bp 16bp 22bp, clip]{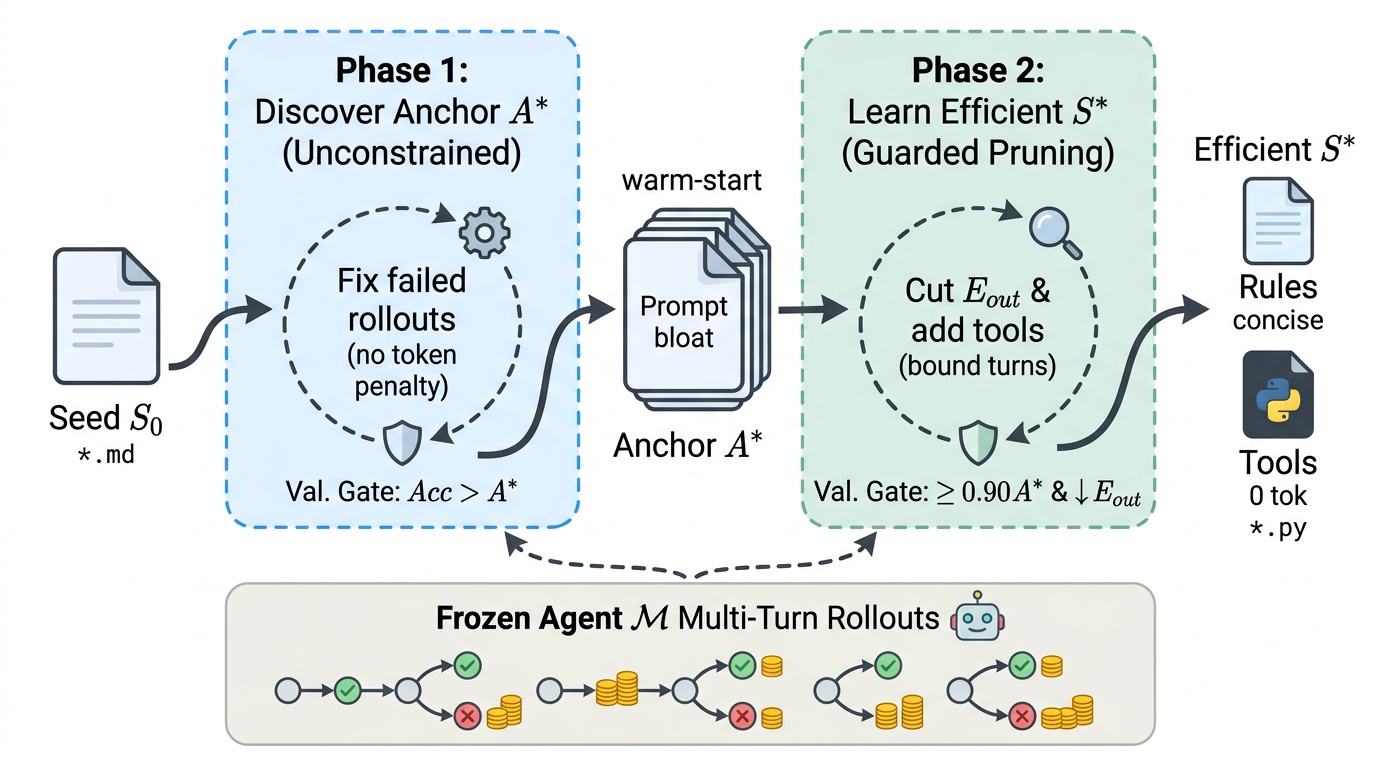}
    \vspace{-2.0mm}
    \caption{\textbf{Overview of \method{}.} From minimal seed $\mathcal{S}_0$, Phase~1 repairs failed rollouts on frozen agent $\mathcal{M}$ without token penalties to discover anchor skill $A^*$. Warm-starting from $A^*$, Phase~2 prunes episodic output tokens ($E_{\text{out}}$) under a $\ge 0.90\,A^*$ validation gate, yielding concise rules and zero-body-cost workspace tools ($\mathcal{S}^*$).}
    \label{fig:overview}
    \vspace{-2.4mm}
\end{figure*}

We introduce \method{} (Figure~\ref{fig:overview}, Algorithm~1), a sequential two-phase framework that \textbf{first discovers a reliable skill ($A^*$)} and \textbf{then learns a token-efficient skill ($\mathcal{S}^*$)}. From minimal seed $\mathcal{S}_0$, Phase~1 repairs failed rollouts without length penalties to establish target-model validation accuracy $A^*$. Phase~2 starts from $A^*$ and evolves to minimize output tokens while ensuring that candidate accuracy is only allowed to be slightly below validation accuracy $A^*$.

Evaluating \method{} across four frontier, mid-tier, and open-weight models on \textsc{MATH-500}, \textsc{ALFWorld}, and \textsc{SpreadsheetBench} against unconstrained skill evolution (\evoskill{}) \citep{evoskill2026} yields four main findings:
\begin{enumerate}[leftmargin=1.3em, itemsep=1.2pt, topsep=1.5pt]
    \item \textbf{Substantial Output Compression (\S\ref{sec:res_compression}):} Across \textsc{MATH-500}, \textsc{ALFWorld}, and \textsc{SpreadsheetBench}, \method{} cuts output tokens on $11$ of $12$ pairs ($38$--$80\%$ on $9$ of $12$), while preserving or improving accuracy.
    \item \textbf{\method{} Acts as Regularization (\S\ref{sec:res_regularization}):} Enforcing token efficiency trims bloated rules and ensures skills are generalized. On \textsc{SpreadsheetBench}, it raises held-out accuracy by $+14.3$~points on Gemini 3.8 Flash and $+17.5$~points on Grok 4.1 Fast by curbing timeouts and early exits.
    \item \textbf{Why the Starting Point Matters (\S\ref{sec:res_coldstart}):} Pruning directly from seed $\mathcal{S}_0$ stalls on unfamiliar tasks since it struggles to find the successful trajectories required to learn the skills in the first place.
    \item \textbf{Lower Serving Cost (\S\ref{sec:res_economics}):} Shorter horizons and offloaded utilities cut billed input tokens by $8$--$66\%$ and lower serving cost across all $12$ pairs ($25$--$67\%$ uncached and $31$--$75\%$ under full-prefix provider KV caching on $10$ of $12$).
\end{enumerate}

\section{Related Work}
\label{sec:related_work}

\paragraph{Agent Skills \& Workspace Tool Libraries.}
Early agent memory and tool-learning systems accumulate trajectories or reusable functions \citep{shinn2023reflexion,wang2023voyager,qian2023creator,cai2024latm}, while TroVE \citep{wang2024trove} and AWM \citep{wang2025awm} induce code toolboxes and workflows. Recent work standardizes modular skill packages pairing Markdown instructions with auxiliary scripts: SkillsBench \citep{li2026skillsbench} finds that focused multi-module skills outperform exhaustive manuals, whereas \evoskill{} \citep{evoskill2026}, GRASP \citep{moll2026grasp}, SkillOpt \citep{skillopt2026}, and SkillLens \citep{huang2026skilllens} evolve skills from failures solely for accuracy. SkillReducer \citep{gao2026skillreducer} trims static instructions ($-$27\% input tokens) without reducing output ($+$1.7\%). In contrast, \method{} jointly optimizes prompt instructions and zero-prompt-cost workspace utilities to cut multi-turn output tokens and turns.

\paragraph{Prompt Optimization \& Reasoning Efficiency.}
Prompt and trace optimizers such as DSPy \citep{khattab2023dspy}, MIPROv2 \citep{opsahl2024mipro}, TextGrad \citep{yuksekgonul2024textgrad}, Trace \citep{cheng2024trace}, and GEPA \citep{gepa2025} refine instructions over multi-step programs and agent rollouts, yet accuracy-only search accumulates long prompts without bounding generation cost. Compile-time structural compression \citep{schnabel2024sammo} shrinks static prompts rather than multi-turn trajectories, whereas brevity prompts \citep{nayab2024concise,chen2025overthinking,han2025tale,xu2025cod} and single-stage cost-regularized optimizers \citep{paretoprompt2025,shah2026crop} penalize generation from the start of search, suppressing task learning when initial accuracy is low. \method{} avoids this cold-start failure via unconstrained discovery followed by accuracy-guarded pruning.

\paragraph{Multi-Turn Reliability \& Context Cost.}
Multi-turn accuracy degrades over long horizons as early mistakes and tracebacks accumulate in context \citep{wang2024mint,yang2024sweagent,cemri2025mast,laban2025lost,khanal2026beyond}. Stateless APIs also replay history at $\mathcal{O}(T^2)$ input cost \citep{ali2026harness,pan2026nlah}, while mid-episode context compression \citep{jiang2023llmlingua,pan2024llmlingua2} can drop critical state and trigger recovery turns \citep{liu2026contextcost}. Like codified workflows \citep{yang2025codeagents}, \method{} improves both cost and reliability at the source via concise instructions and workspace utilities.

\section{\method{}}
\label{sec:method}

\paragraph{Skill Representation \& Token Cost.}
\label{sec:package}
An agent skill $\mathcal{S} = (\mathcal{I}, \mathcal{T})$ pairs \textbf{prompt instructions} $\mathcal{I}$ (natural-language guidance supplied in context) with \textbf{workspace utilities} $\mathcal{T}$ (reusable executable functions placed in the agent's working directory). Offloading function bodies into $\mathcal{T}$ eliminates their system-prompt token cost, leaving only compact call signatures in $\mathcal{I}$ while any runtime output enters observation history. Given a task, a frozen agent $\mathcal{M}$ conditioned on $\mathcal{S}$ interacts over $T$ turns to produce $\tau = (o_1, a_1, \dots, o_T, a_T)$ with episodic output tokens $E_{\text{out}}(\tau)$ across internal reasoning and visible actions. Concise instructions $\mathcal{I}$ and offloaded utilities $\mathcal{T}$ curb both turn count $T$ and output tokens $E_{\text{out}}(\tau)$.

\begin{figure}[t]
\phantomsection\label{alg:efficientskill}
\footnotesize
\renewcommand{\arraystretch}{1.04}
\newcommand{\algln}[1]{\makebox[1.15em][r]{\scriptsize\textcolor{inkMuted}{#1:}}\hspace{0.4em}}
\begin{tabular}{@{}>{\raggedright\arraybackslash}p{\columnwidth}@{}}
\toprule
\textbf{Algorithm 1:} \method{} \\
\midrule
\textbf{Input:} Agent $\mathcal{M}$, seed $\mathcal{S}_0 = (\mathcal{I}_0, \varnothing)$, $\mathcal{D}_{\text{train}}, \mathcal{D}_{\text{val}}$, $\alpha, \delta$ \\
\textbf{Output:} Token-efficient skill $\mathcal{S}^* = (\mathcal{I}^*, \mathcal{T}^*)$ \\
\addlinespace[1.2pt]
\textcolor{codeKw}{\textbf{Phase 1: Discover Anchor $A^*$}} \hfill {\scriptsize\textcolor{inkMuted}{// unconstrained}} \\
\algln{1}$\mathcal{S}_{\text{acc}}, \mathcal{S}' \leftarrow \mathcal{S}_0;\; A^* \leftarrow A(\mathcal{S}_0)$ \\
\algln{2}\textbf{for} mini-batch $\mathcal{B} \subset \mathcal{D}_{\text{train}}$ \textbf{until} patience \textbf{do} \\
\algln{3}\hspace{0.7em}$\mathcal{S}' \leftarrow \textsc{Repair}(\mathcal{S}', \textsc{Rollout}(\mathcal{M}, \mathcal{S}', \mathcal{B}))$ \hfill {\scriptsize\textcolor{inkMuted}{// fix failures}} \\
\algln{4}\hspace{0.7em}\textbf{at ckpt:} \textbf{if} $A(\mathcal{S}') > A^*$ \textbf{then} $\mathcal{S}_{\text{acc}}, A^* \leftarrow \mathcal{S}', A(\mathcal{S}')$ \\
\algln{5}\hspace{3.65em}\textbf{else} $\mathcal{S}' \leftarrow \mathcal{S}_{\text{acc}}$ \\
\addlinespace[1.2pt]
\textcolor{gainForest}{\textbf{Phase 2: Learn Efficient $\mathcal{S}^*$}} \hfill {\scriptsize\textcolor{inkMuted}{// warm-start from $\mathcal{S}_{\text{acc}}$}} \\
\algln{6}$\mathcal{S}^*, \mathcal{S}' \leftarrow \mathcal{S}_{\text{acc}};\; E_{\text{best}} \leftarrow E_{\text{out}}(\mathcal{S}_{\text{acc}})$ \\
\algln{7}\textbf{for} mini-batch $\mathcal{B} \subset \mathcal{D}_{\text{train}}$ \textbf{until} patience \textbf{do} \\
\algln{8}\hspace{0.7em}$\mathcal{S}' \leftarrow \textsc{Prune}(\mathcal{S}', \textsc{Rollout}(\mathcal{M}, \mathcal{S}', \mathcal{B}))$ \hfill {\scriptsize\textcolor{inkMuted}{// cut $\mathcal{I}'$, add $\mathcal{T}'$}} \\
\algln{9}\hspace{0.7em}\textbf{at ckpt:} \textbf{if} $A(\mathcal{S}') \ge \alpha A^*$ \textbf{and} $E_{\text{out}}(\mathcal{S}') \le (1{-}\delta)E_{\text{best}}^\dagger$ \\
\algln{10}\hspace{3.65em}\textbf{then} $\mathcal{S}^*, E_{\text{best}} \leftarrow \mathcal{S}', E_{\text{out}}(\mathcal{S}')$ \textbf{else} $\mathcal{S}' \leftarrow \mathcal{S}^*$ \\
\algln{11}\textbf{return} $\mathcal{S}^*$ \\
\midrule
{\scriptsize\textcolor{inkMuted}{$^\dagger$Requires valid $\mathcal{T}'$; also commits Pareto/acc.\ gains ${\le}1.1 E_{\text{best}}$ (App.~\ref{sec:app_opt_schedule}).}} \\
\bottomrule
\end{tabular}
\vspace{-3.4mm}
\end{figure}

\paragraph{Sequential Optimization.}
\label{sec:two_phase}
Let $A(\mathcal{S})$ and $E_{\text{out}}(\mathcal{S})$ denote validation pass rate and mean episodic output tokens on $\mathcal{D}_{\text{val}}$. After each training mini-batch, a reflective meta-optimizer proposes provisional edits $\mathcal{S}' = (\mathcal{I}', \mathcal{T}')$, evaluated on $\mathcal{D}_{\text{val}}$ at periodic checkpoints ($4$/epoch) and committed or reverted to the incumbent (Appendix~\ref{sec:app_opt_schedule}). Because penalizing tokens directly from $\mathcal{S}_0 = (\mathcal{I}_0, \varnothing)$ suppresses early task learning, \method{} runs two sequential phases (Figure~\ref{fig:overview}, Algorithm~1):
\begin{enumerate}[leftmargin=1.35em, itemsep=1.2pt, topsep=1.5pt]
    \item \textbf{Discover Optimized Skill $A^*$ (Unconstrained):} Starting from $\mathcal{S}_0$, Phase~1 ($\textsc{Repair}$) fixes failed rollouts without token penalties, committing improvements ($A(\mathcal{S}') > A^*$) to establish anchor skill $\mathcal{S}_{\text{acc}}$ with validation accuracy $A^*$.
    \item \textbf{Learn Efficient $\mathcal{S}^*$:} Warm-starting $(A_{\text{best}}, E_{\text{best}}) \leftarrow (A^*, E_{\text{out}}(\mathcal{S}_{\text{acc}}))$ under fixed discrete floor $\alpha A^*$ (Appendix~\ref{sec:app_opt_schedule}), Phase~2 ($\textsc{Prune}$) tightens prose, bounds turns, and offloads code into compilation-checked $\mathcal{T}'$. Checkpoints commit $\mathcal{S}'$ under $A(\mathcal{S}') \ge \alpha A^*$ when $E_{\text{out}}$ drops by $\ge \delta$ or Pareto-improves $(A_{\text{best}}, E_{\text{best}})$.
\end{enumerate}

\section{Experimental Setup}
\label{sec:setup}

\begin{table*}[t]
\centering
\footnotesize
\renewcommand{\arraystretch}{1.04}
\caption{Held-out test performance ($N_{\text{test}}=280$) comparing \evoskill{} against \method{} in task pass rate (\%), mean turns ($\bar{T}$), and mean output tokens per episode ($E_{\text{out}}$, including reasoning).}
\label{tab:main_results}
\begin{tabular*}{\textwidth}{@{\extracolsep{\fill}}l rr rr rrr@{}}
\toprule
& \multicolumn{2}{c}{\textbf{Test Pass (\%)}} & \multicolumn{2}{c}{\textbf{Mean Turns ($\bar{T}$)}} & \multicolumn{3}{c}{\textbf{Output Tokens / Episode ($E_{\text{out}}$)}} \\
\cmidrule(lr){2-3} \cmidrule(lr){4-5} \cmidrule(lr){6-8}
\textbf{Model} & \evoskill{} & \textbf{Ours} & \evoskill{} & \textbf{Ours} & \evoskill{} & \textbf{Ours} & $\Delta_{\text{out}}$ \\
\midrule
\multicolumn{8}{@{}l}{\textbf{\textsc{MATH-500}} \textnormal{\scriptsize(Competition Math with Optional Python)}} \\
\hspace{0.4em}Claude Opus 5.5   & 86.4 & \textbf{87.5} & 1.00 & 1.00          & 426    & \textbf{102}    & \textbf{$-$76.0\%} \\
\hspace{0.4em}Gemini 3.8 Flash  & 85.0 & \textbf{87.5} & 1.76 & \textbf{1.04} & 1,734  & \textbf{543}    & \textbf{$-$68.7\%} \\
\hspace{0.4em}Grok 4.1 Fast     & 72.5 & 72.5          & 1.01 & \textbf{1.00} & 1,937  & \textbf{1,665}  & \textbf{$-$14.1\%} \\
\hspace{0.4em}Gemma 4 26B       & 80.4 & \textbf{83.6} & 1.70 & \textbf{1.19} & 1,349  & \textbf{806}    & \textbf{$-$40.3\%} \\
\midrule
\multicolumn{8}{@{}l}{\textbf{\textsc{ALFWorld}} \textnormal{\scriptsize(Embodied Household Navigation)}} \\
\hspace{0.4em}Claude Opus 5.5   & 88.9          & \textbf{93.2} & 10.21 & \textbf{7.33} & 933    & \textbf{376}    & \textbf{$-$59.7\%} \\
\hspace{0.4em}Gemini 3.8 Flash  & 92.1          & \textbf{93.6} & 9.85  & \textbf{7.16} & 5,704  & \textbf{1,155}  & \textbf{$-$79.7\%} \\
\hspace{0.4em}Grok 4.1 Fast     & \textbf{89.3} & 88.2          & 9.15  & \textbf{8.71} & 12,761 & \textbf{12,350} & \textbf{$-$3.2\%} \\
\hspace{0.4em}Gemma 4 26B       & \textbf{81.4} & 81.1          & 10.74 & \textbf{9.12} & \textbf{95} & 99         & $+$4.4\%$^\S$ \\
\midrule
\multicolumn{8}{@{}l}{\textbf{\textsc{SpreadsheetBench}} \textnormal{\scriptsize(Multi-Turn Programmatic Workbook Editing)}} \\
\hspace{0.4em}Claude Opus 5.5   & 88.6          & 88.6          & \textbf{2.80} & 3.04  & 1,466  & \textbf{761}    & \textbf{$-$48.1\%} \\
\hspace{0.4em}Gemini 3.8 Flash  & 58.2          & \textbf{72.5} & 10.44 & \textbf{8.56} & 13,376 & \textbf{8,364}  & \textbf{$-$37.5\%} \\
\hspace{0.4em}Grok 4.1 Fast     & 57.5          & \textbf{75.0} & 2.71  & \textbf{1.97} & 10,056 & \textbf{4,271}  & \textbf{$-$57.5\%} \\
\hspace{0.4em}Gemma 4 26B       & \textbf{47.1} & 45.7          & 5.41  & \textbf{3.39} & 4,677  & \textbf{2,174}  & \textbf{$-$53.5\%} \\
\bottomrule
\end{tabular*}
\par\vspace{0.8mm}
\begin{minipage}{0.995\textwidth}
\scriptsize
\emph{Notes:} Bold marks strictly better values per pair ($\Delta_{\text{out}}$ from unrounded totals; $1$ test task $= 0.36$ pts, binomial $\text{SE} \le 3.0$ pts; validation progression and serving costs in Tables~\ref{tab:economics_roi} and \ref{tab:app_feasibility_gate}). Under a $5\%$-pt slack floor, Gemma reaches $47.9\%$ on \textsc{SpreadsheetBench} ($-$57.9\% OutTok). $^\S$On \textsc{ALFWorld}, Gemma emits $0$ reasoning tokens ($95.2 \to 99.4$ action tokens), while fewer turns cut billed input tokens by $36.9\%$ and serving cost by $36.7\%$ (Table~\ref{tab:economics_roi}).
\end{minipage}
\vspace{-2.5mm}
\end{table*}

\paragraph{Benchmarks \& Verifiers.}
We evaluate three interactive benchmarks with deterministic binary verifiers ($\mathcal{R} \in \{0,1\}$; Appendix~\ref{sec:appendix_benchmarks}):
\begin{itemize}[leftmargin=1.2em, itemsep=1.5pt, topsep=2pt]
    \item \textbf{\textsc{MATH-500}} \citep{hendrycksmath2021}: Competition mathematics with optional Python execution, graded by symbolic equivalence.
    \item \textbf{\textsc{ALFWorld}} \citep{alfworld2020}: Multi-turn household navigation and object manipulation, graded by simulator goal predicates.
    \item \textbf{\textsc{SpreadsheetBench}} \citep{spreadsheetbench2024}: Real-world spreadsheet editing via Python scripts, graded by cell-level value matching against reference workbooks.
\end{itemize}

\paragraph{Models \& Evaluation Protocol.}
All methods run in a shared stateless execution environment with isolated per-episode workspaces and exact per-turn token logging (Appendix~\ref{sec:appendix_benchmarks}). Benchmarks use frozen $20\%/10\%/70\%$ train/validation/test splits. Each selected skill is evaluated once on $\mathcal{D}_{\text{test}}$ across four models: \textbf{Claude Opus 5.5}, \textbf{Gemini 3.8 Flash}, \textbf{Grok 4.1 Fast}, and open-weight \textbf{Gemma 4 26B}.

\paragraph{Optimization Schedule \& Baseline.}
We instantiate $\mathcal{S} = (\mathcal{I}, \mathcal{T})$ with empty Markdown instructions (\texttt{SKILL.md}; Appendix~\ref{sec:app_seed_prompts}) and a Python module (\texttt{tools.py}), using Claude Opus 5.5 as meta-optimizer. Gemini Argon as meta-optimizer showed similar performance. Each phase runs mini-batches of $B=8$ training tasks for up to $10$ epochs ($4$ validation checkpoints/epoch, patience $6$, $\alpha=0.90$, $\delta=0.05$; Appendix~\ref{sec:app_opt_schedule}). \evoskill{} \citep{evoskill2026} runs unconstrained accuracy-only search under Phase~1's identical schedule ($\mathcal{S}_{\text{acc}}$); Phase~2 adds warm-started pruning (extended-budget \evoskill{} control in Appendix~\ref{sec:app_ablation_budget}; marginal spend in Table~\ref{tab:roi_breakeven}).

\section{Results}
\label{sec:results}

\subsection{Output-Token Compression}
\label{sec:res_compression}

Table~\ref{tab:main_results} reports held-out test performance across all $12$ model--benchmark pairs. Across all three benchmarks, \method{} cuts output generation on $11$ of $12$ pairs while raising mean benchmark accuracy:
\begin{itemize}[leftmargin=1.2em, itemsep=1.0pt, topsep=1.2pt]
    \item \textbf{\textsc{MATH-500}:} \method{} cuts output tokens by $40$--$76\%$ on Claude Opus 5.5, Gemini 3.8 Flash, and Gemma 4 26B ($14\%$ on Grok 4.1 Fast) while matching or slightly exceeding \evoskill{} accuracy. Learned skills replace multi-turn code retries with direct symbolic derivation, cutting turns from $1.76 \to 1.04$ on Gemini and $1.70 \to 1.19$ on Gemma (App.~\ref{sec:app_math500_derivation}).
    \item \textbf{\textsc{ALFWorld}:} On Opus and Gemini, where output is dominated by deliberation, distilling compact object-location priors cuts output tokens by $60\%$ and $80\%$ and lifts Opus accuracy by $+4.3$~points ($88.9\% \to 93.2\%$). On Gemma 4 ($0$ reasoning tokens), finishing in fewer turns ($10.74 \to 9.12$) still cuts billed input tokens and cost by $37\%$ (Table~\ref{tab:economics_roi}).
    \item \textbf{\textsc{SpreadsheetBench}:} \method{} cuts output tokens by $38$--$58\%$ across all four models ($48\%$ on Opus, $38\%$ on Gemini, $58\%$ on Grok, and $54\%$ on Gemma) by shortening multi-turn editing loops and moving recurring code into workspace utilities.
\end{itemize}

\begin{figure}[t]
    \centering
    \includegraphics[width=\columnwidth]{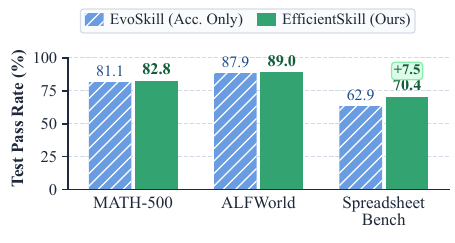}
    \vspace{-5.2mm}
    \caption{\textbf{Pruning Acts as Regularization.} Mean held-out pass rate across all four models improves on every benchmark under \method{}, led by $+7.5$~points on \textsc{SpreadsheetBench} ($+7.6$ unrounded).}
    \label{fig:regularization}
    \vspace{-3.2mm}
\end{figure}

\subsection{Pruning Acts as Regularization, Leading to Better Performance}
\label{sec:res_regularization}

Across all three benchmarks, guarded pruning raises four-model mean test accuracy ($81.1\% \to 82.8\%$ on \textsc{MATH-500}, $87.9\% \to 89.0\%$ on \textsc{ALFWorld}, and $62.9\% \to 70.4\%$ on \textsc{SpreadsheetBench}; Figure~\ref{fig:regularization}). Gains are largest on \textsc{SpreadsheetBench}, where unconstrained \evoskill{} expands prompts to $908$--$1{,}521$ words with empty \texttt{tools.py} modules (Appendix~\ref{sec:app_ablation_budget}):
\begin{itemize}[leftmargin=1.2em, itemsep=1.0pt, topsep=1.2pt]
    \item \textbf{Fewer Timeouts and Early Exits:} \method{} lifts test accuracy by $+14.3$~points on Gemini 3.8 Flash ($58.2\% \to 72.5\%$) and $+17.5$~points on Grok 4.1 Fast ($57.5\% \to 75.0\%$) while matching Opus ($88.6\%$; $-1.4$~points on Gemma), cutting combined timeout and deliberation-drift failures on Gemini ($35.7\% \to 19.6\%$; $20$-turn timeouts $16.1\% \to 6.4\%$) and premature Turn-1 exits on Grok ($18.2\% \to 3.2\%$).
    \item \textbf{Shorter Prompts and Reusable Helpers:} Phase~2 acts jointly on both components, shrinking prompt instructions by $23$--$59\%$ (Table~\ref{tab:economics_roi}) while populating $241$--$554$ words of reusable inspection, parsing, and verification helpers in \texttt{tools.py} (Table~\ref{tab:app_search_budget}; Appendix~\ref{sec:app_skill_gemini_sheet}).
\end{itemize}

\begin{figure}[t]
    \centering
    \includegraphics[width=\columnwidth]{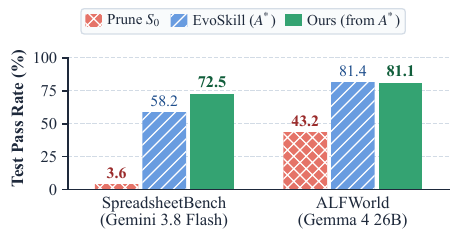}
    \vspace{-5.2mm}
    \caption{\textbf{Starting Point Matters.} Pruning directly from seed $\mathcal{S}_0$ stalls and reverts ($N=280$), whereas warm-starting from $A^*$ (\method{}) reaches $72.5\%$ and $81.1\%$.}
    \label{fig:coldstart}
    \vspace{-3.2mm}
\end{figure}

\subsection{Why the Starting Point for Efficiency Optimization Matters}
\label{sec:res_coldstart}

What if Phase~2's pruning prompt and static $0.90\,A^*$ gate are applied directly from $\mathcal{S}_0$ without Phase~1 warm-starting? In Figure~\ref{fig:coldstart} (Appendix~\ref{sec:app_cold_start_trap}), early length pressure suppresses exploratory task rules: even the best candidate proposals prior to gate rejection stall at $20.0\%$ on \textsc{SpreadsheetBench} (vs.\ $50.0\%$ unconstrained) and $71.8\%$ on \textsc{ALFWorld} (vs.\ $89.7\%$), failing the $0.90\,A^*$ gate until patience reverts to $\mathcal{S}_0$ ($3.6\%$ and $43.2\%$ test pass, even at $B=24$, vs.\ $72.5\%$ and $81.1\%$ warm-started).

\begin{table*}[t]
\centering
\footnotesize
\setlength{\tabcolsep}{5.0pt}
\renewcommand{\arraystretch}{1.02}
\caption{\texttt{SKILL.md} word counts (\evoskill{} $\to$ \method{}), billed input-token change, and standard serving-cost change across all $12$ held-out test configurations ($N_{\text{test}}=280$).}
\label{tab:economics_roi}
\begin{tabular*}{\textwidth}{@{\extracolsep{\fill}}l c r r@{}}
\toprule
\textbf{Model} & \textbf{\texttt{SKILL.md} Words ($\evoskill{} \to \textbf{Ours}$)} & \textbf{Input Tok.\ Change} & \textbf{Serving Cost Change} \\
\midrule
\multicolumn{4}{@{}l}{\textbf{\textsc{MATH-500}} \textnormal{\scriptsize(Direct math derivation $+$ fewer code retries)}} \\
\hspace{0.4em}Claude Opus 5.5   & $1{,}157 \to \mathbf{649}$ & $-$36.1\% & \textbf{$-$51.9\%} \\
\hspace{0.4em}Gemini 3.8 Flash  & $1{,}073 \to \mathbf{863}$ & $-$56.2\% & \textbf{$-$63.9\%} \\
\hspace{0.4em}Grok 4.1 Fast     & $1{,}111 \to \mathbf{781}$ & $-$8.4\%  & \textbf{$-$12.1\%} \\
\hspace{0.4em}Gemma 4 26B       & $1{,}134 \to \mathbf{757}$ & $-$58.2\% & \textbf{$-$55.1\%} \\
\midrule
\multicolumn{4}{@{}l}{\textbf{\textsc{ALFWorld}} \textnormal{\scriptsize(Compact search rules $+$ fewer navigation steps)}} \\
\hspace{0.4em}Claude Opus 5.5   & $1{,}175 \to \mathbf{500}$ & $-$65.3\% & \textbf{$-$64.6\%} \\
\hspace{0.4em}Gemini 3.8 Flash  & $1{,}135 \to \mathbf{656}$ & $-$54.6\% & \textbf{$-$67.3\%} \\
\hspace{0.4em}Grok 4.1 Fast     & $1{,}265 \to \mathbf{753}$ & $-$29.9\% & \textbf{$-$15.2\%} \\
\hspace{0.4em}Gemma 4 26B       & $1{,}341 \to \mathbf{880}$ & $-$36.9\% & \textbf{$-$36.7\%} \\
\midrule
\multicolumn{4}{@{}l}{\textbf{\textsc{SpreadsheetBench}} \textnormal{\scriptsize(Reusable helper functions $+$ fewer editing turns)}} \\
\hspace{0.4em}Claude Opus 5.5   & $1{,}521 \to \mathbf{785}$ & $-$25.5\% & \textbf{$-$32.8\%} \\
\hspace{0.4em}Gemini 3.8 Flash  & $1{,}271 \to \mathbf{521}$ & $-$17.0\% & \textbf{$-$25.3\%} \\
\hspace{0.4em}Grok 4.1 Fast     & $908 \to \mathbf{703}$     & $-$45.6\% & \textbf{$-$54.3\%} \\
\hspace{0.4em}Gemma 4 26B       & $1{,}198 \to \mathbf{698}$ & $-$65.6\% & \textbf{$-$64.2\%} \\
\bottomrule
\end{tabular*}
\par\vspace{0.6mm}
\begin{minipage}{0.995\textwidth}
\scriptsize
\emph{Notes:} Full-prefix provider KV-caching costs ($31$--$75\%$ savings on $10/12$ pairs), \texttt{tools.py} word counts, and Phase~2 optimization payback appear in Appendices~\ref{sec:app_ablation_budget}--\ref{sec:app_input_context_economics} (Tables~\ref{tab:app_search_budget}, \ref{tab:input_context_economics}, and \ref{tab:roi_breakeven}).
\end{minipage}
\vspace{-3.2mm}
\end{table*}

\subsection{Input Tokens \& Serving Cost}
\label{sec:res_economics}

Table~\ref{tab:economics_roi} summarizes skill instruction compression, billed input-token reductions, and serving-cost savings across all $12$ pairs (Appendix~\ref{sec:app_input_context_economics}):
\begin{itemize}[leftmargin=1.2em, itemsep=1.0pt, topsep=1.2pt]
    \item \textbf{Lower Input Tokens and Serving Cost:} Shorter skill instructions ($20$--$59\%$ fewer words in \texttt{SKILL.md}), offloaded \texttt{tools.py} helpers, and fewer turns per trajectory cut cumulative input tokens by $8$--$66\%$ and lower standard serving cost across all $12$ pairs ($25$--$67\%$ on $10$ of $12$).
    \item \textbf{Savings Hold Under Prefix Caching:} Even when discounting reused conversation prefixes at provider KV-cache rates (Appendix~\ref{sec:app_input_context_economics}, Table~\ref{tab:input_context_economics}), serving cost still drops on all $12$ pairs ($31$--$75\%$ on $10$ of $12$) because pricier output tokens and newly appended turn observations remain uncached (Phase~2 payback in Table~\ref{tab:roi_breakeven}).
\end{itemize}

\section{Trajectory Analysis}
\label{sec:diagnostics}
\label{sec:forensics}

Paired test traces clarify how guarded pruning reshapes multi-turn behavior (Appendix~\ref{sec:appendix_forensics}):
\begin{itemize}[leftmargin=1.2em, itemsep=1.0pt, topsep=1.2pt]
    \item \textbf{Stopping No-Code Loops (Gemini 3.8 Flash):} Under \evoskill{} ($1{,}271$-word prompt, empty \texttt{tools.py}), once Turn~1--2 inspection dumps cell layouts, reconciling long prose rules without callable utilities leaves $26.1\%$ of turns ($763/2{,}923$) without executable code. Unexecuted turns leave the file unchanged in context, triggering self-reinforcing no-code loops ($27\%$ follow another no-code turn) and compounding with inline scripting errors ($5.8\% \to 10.4\%$ next-turn error rate). \method{} pairs concise rules with working utilities, cutting total unexecuted turns to $14.1\%$ ($338/2{,}398$, mostly code-free completion turns; non-empty prose turns fall to $1.1\%$, $27/2{,}398$) and $20$-turn timeouts by $60\%$ ($45 \to 18$).
    \item \textbf{Preventing Unsaved Submissions (Grok 4.1 Fast):} Under \evoskill{}, verbose Turn-1 plans often emit the completion signal before saving edits, ending $51/280$ runs ($18.2\%$, $0\%$ pass) on untouched files. \method{} requires verified edits first, cutting early exits to $9/280$ ($3.2\%$) and recovering $40$ of those $51$ tasks ($+49$-task net gain).
    \item \textbf{Solving Tasks in Fewer Steps:} Rather than piecemeal probing, \method{} converges on \textsc{SpreadsheetBench} to an \emph{inspect $\to$ compute-and-verify $\to$ submit} loop ($93.2\%$ of Opus runs finish in $3$ turns) backed by verification helpers. Across non-Opus \textsc{SpreadsheetBench} models, this cuts both turns ($18$--$37\%$) and tokens per turn ($24$--$42\%$), while Opus halves per-turn output ($-52\%$; App.~\ref{sec:app_horizon_decomposition}).
\end{itemize}

\section{Practical Takeaways}
\label{sec:playbook}

\begin{enumerate}[leftmargin=1.3em, itemsep=1.1pt, topsep=1.4pt]
    \item \textbf{Get the skill working before compressing:} First run unconstrained repair to reach validation accuracy $A^*$, then prune under $\ge 0.90\,A^*$.
    \item \textbf{Move repeated code into helper scripts:} On programmatic tasks, moving deterministic inspection, parsing, and verification code into workspace modules avoids prompt-body bloat and curbs inline scripting errors.
    \item \textbf{Cap output tokens to curb multi-turn drift:} Bounding episodic output enforces compact multi-turn workflows, curbing timeout spirals, uncached output generation, and residual cached-read history costs.
    \item \textbf{Re-optimize for each target model:} Because learned skills adapt to each model's verbosity and error profile at modest Phase~2 cost, re-run guarded pruning when switching models.
\end{enumerate}

\section{Conclusion}
\label{sec:conclusion}

When skill optimizers chase accuracy alone, they keep patching edge cases until over-fitted skills pull agents into hesitant deliberation and costly retries. Yet penalizing token use too early backfires, because learning an unfamiliar task initially requires both success and failure trajectories to see what works well. \method{} resolves this in two phases: Phase~1 learns a working skill without token limits, and Phase~2 refines that skill and its helper tools to produce shorter, token-efficient trajectories while keeping validation accuracy near its peak. Across four models and three benchmarks, replacing brittle patches with compact skills and reusable code helps agents generalize, raising mean test accuracy on every benchmark. At the same time, trajectories with fewer turns and less generated output cut serving cost in every setting, even under full-prefix caching, which discounts only re-read history. Optimizing skills for token-efficient execution improves reliability rather than trading it away, once optimization first learns how to solve a task before learning how to solve it concisely.

\clearpage
\section*{Limitations}
\label{sec:limitations}

\begin{itemize}[leftmargin=1.25em, itemsep=2.2pt, topsep=2pt]
    \item \textbf{Optimizing Output Tokens Instead of Dollar Cost:} Guarded pruning gates candidate acceptance on episodic output tokens $E_{\text{out}}$ so learned packages remain pricing-agnostic across providers, rather than optimizing a provider-specific dollar-cost objective. Although output compression and turn reduction lowered both uncached and full-prefix cached serving costs across all $12$ evaluated pairs, fixed-horizon workflows where a candidate substantially expands prompt instructions $|\mathcal{I}|$ could in principle raise uncached input replay cost; gating directly on weighted API spend is a straightforward extension.
    \item \textbf{Meta-Optimizer Cost, Output Cap, and Baseline Scope:} Both phases query Claude Opus 5.5 as the reflective meta-optimizer under a shared $5{,}000$-token output cap. While target-model rollout spend ($C_{\text{prune}}$) is under $\$3.60$ on $8$ of $9$ non-Opus runs, Opus meta-optimizer calls add $\approx\$1.80$--$\$9.36$ per run, dominating offline dollar cost on inexpensive single-turn runs (Table~\ref{tab:roi_breakeven}); moreover, when unconstrained \evoskill{} expands \texttt{SKILL.md} past $900$ words on \textsc{SpreadsheetBench}, the $5{,}000$-token cap truncates trailing \texttt{tools\_py} drafts (though trace audits in Appendix~\ref{sec:app_skill_gemini_sheet} show missing-\texttt{tools} errors affect $\le 1.1$~accuracy points). Finally, Phase~2 jointly compresses \texttt{SKILL.md} and populates \texttt{tools.py}; evaluating larger reflection token budgets, isolated prompt-only pruning (\texttt{tools.py} disabled), post-hoc static compressors \citep{gao2026skillreducer}, and soft running-incumbent regularizers \citep{paretoprompt2025,shah2026crop}, as well as open-ended tasks graded by noisy model judges, remains future work.
\end{itemize}

\bibliography{custom}

\clearpage
\onecolumn
\appendix
\makeatletter\setlength{\@fptop}{0pt}\makeatother
\section{Evaluation Environments, Verifiers \& Hyperparameters}
\label{sec:appendix_benchmarks}
\label{sec:app_pomdp_specs}
\label{sec:app_dataset_splits}

\subsection{Benchmarks, Execution Environment \& Grading}
\label{sec:app_verifiers}

\begin{table}[htbp]
\centering
\footnotesize
\setlength{\tabcolsep}{6pt}
\renewcommand{\arraystretch}{1.14}
\caption{Benchmark specifications, frozen $20\%/10\%/70\%$ dataset splits (seed $42$; $1{,}198$ unique tasks: $400$-task \textsc{SpreadsheetBench}-Verified, first $80/40/280$ of shuffled \textsc{MATH-500} for matched $N_{\text{test}}=280$, and $398$ \textsc{ALFWorld} games balanced across all $6$ task types), action spaces with per-episode turn caps, and deterministic binary verifiers.}
\label{tab:app_pomdp_summary}
\begin{tabularx}{\textwidth}{@{}l c >{\raggedright\arraybackslash}p{4.6cm} >{\raggedright\arraybackslash}X@{}}
\toprule
\textbf{Benchmark} & \textbf{Train / Val / Test} & \textbf{Action Space \& Turn Cap} & \textbf{Deterministic Verifier ($\mathcal{R} \in \{0,1\}$)} \\
\midrule
\textbf{\textsc{MATH-500}} & 80 / 40 / 280 & Optional Python code block or final \texttt{\textbackslash boxed\{\dots\}} (up to $10$ turns) & Exact string match or symbolic equivalence via \texttt{SymPy} ($3$\,s timeout). \\
\addlinespace[3pt]
\textbf{\textsc{ALFWorld}} & 79 / 39 / 280 & One TextWorld command per turn (up to $25$ turns) & Simulator goal check run after every step. \\
\addlinespace[3pt]
\textbf{\textsc{SpreadsheetBench}} & 80 / 40 / 280 & Python script (\texttt{openpyxl}) editing target workbook, or completion signal (up to $20$ turns) & Cell-by-cell cached-value match against reference workbook ($|v_a - v_g| < 10^{-4}$ for floats; case-insensitive trimmed match for strings). \\
\bottomrule
\end{tabularx}
\end{table}

\begin{itemize}[leftmargin=1.4em, itemsep=2pt, topsep=3pt]
    \item \textbf{Stateless Evaluation Environment:} Table~\ref{tab:app_pomdp_summary} summarizes all three benchmarks ($10$, $25$, and $20$ maximum turns per episode on \textsc{MATH-500}, \textsc{ALFWorld}, and \textsc{SpreadsheetBench}). Rather than wrapping vendor-specific CLI agents (such as Claude Code or Codex), we run all four models in the same stateless multi-turn loop. Skill instructions (\texttt{SKILL.md}) pass as the system prompt, while helper scripts (\texttt{tools.py}) sit on disk in the episode workspace; full function bodies never enter the system prompt, whereas short call signatures in \texttt{SKILL.md} and printed runtime output count toward Turn-1 and cumulative input tokens ($I_1$ and $E_{\text{in}}$; API and cache pricing in Table~\ref{tab:input_context_economics}). Each turn resends the prior action and observation history (without past internal reasoning tokens, which providers strip automatically) and logs exact input, reasoning, and completion tokens. Code benchmarks extract fenced Python blocks from model replies, while \textsc{ALFWorld} extracts a single TextWorld command.
    \item \textbf{Isolated Code Sandboxes:} Each \textsc{MATH-500} and \textsc{SpreadsheetBench} episode runs in a fresh temporary working directory with standard input disabled. Subprocesses exceeding the per-turn timeout ($10$\,s on \textsc{MATH-500}; $15$\,s on \textsc{SpreadsheetBench}) are killed, and standard output and error streams are each capped at $4{,}000$ and $8{,}000$ characters, respectively.
    \item \textbf{\textsc{SpreadsheetBench} Grading:} Each episode places the initial workbook at \texttt{input.xlsx} and instructs the agent to edit and save \texttt{input.xlsx} in place before outputting \texttt{SUBMIT: DONE}. At submission, the grader opens \texttt{input.xlsx} via \texttt{openpyxl} (\texttt{data\_only=True}) without running Excel or LibreOffice. Newly written formula strings carry no cached value and read back as \texttt{None}, and edits saved only under another filename (such as \texttt{output.xlsx}) leave \texttt{input.xlsx} unchanged.
\end{itemize}

\subsection{Optimization Schedule \& State Updates}
\label{sec:app_opt_schedule}
\begin{itemize}[leftmargin=1.4em, itemsep=2pt, topsep=3pt]
    \item \textbf{Mini-Batch Schedule \& Validation Checkpoints:} Both phases (Algorithm~1, \S\ref{sec:method}) train on mini-batches of $B=8$ tasks for up to $\mathcal{E}=10$ epochs ($10$ batches per $80$-task epoch; on \textsc{ALFWorld}, batch~$10$ has $7$ tasks). Four validation checkpoints run per epoch after batch indices $b \in \{3, 5, 8, 10\}$ ($\lceil 10k/4 \rceil$ for $k=1,\dots,4$, spaced every $3, 2, 3, 2$ batches), with early-stopping patience $\mathcal{P}=6$ checkpoints ($15$ training batches). After every batch, the Claude Opus 5.5 meta-optimizer ($5{,}000$ max output tokens; Appendix~\ref{sec:appendix_prompts}) proposes a JSON edit to the working skill $\mathcal{S}'$ without receiving prior rejection history; episodes with transient API or network errors are dropped before reflection. Between checkpoints ($b \notin \{3, 5, 8, 10\}$), $\mathcal{S}'$ carries forward directly to the next training batch. At checkpoints, accepted candidates update the incumbent ($\mathcal{S}_{\text{acc}}$ or $\mathcal{S}^*$) and reset patience to $0$, whereas rejected candidates roll $\mathcal{S}'$ back to the incumbent and add $1$ to the patience counter.
    \item \textbf{Code Compilation Check \& Checkpoint Acceptance Rules:} Whenever a proposed edit includes non-empty \texttt{tools.py} code, syntax and bytecode compilation are checked in an isolated subprocess ($5$\,s timeout), and invalid \texttt{tools.py} drafts are discarded before evaluation. Phase~1 (and unconstrained \evoskill{}) commits $\mathcal{S}'$ at a checkpoint when $A(\mathcal{S}') > A^*$ (or $\ge A(\mathcal{S}_0)$ while still at the initial seed $\mathcal{S}_0$), updating $(\mathcal{S}_{\text{acc}}, A^*) \leftarrow (\mathcal{S}', A(\mathcal{S}'))$. Phase~2 warm-starts $(\mathcal{S}^*, A_{\text{best}}, E_{\text{best}}) \leftarrow (\mathcal{S}_{\text{acc}}, A^*, E_{\text{out}}(\mathcal{S}_{\text{acc}}))$ and sets a fixed discrete accuracy floor $A_{\text{floor}} = A^* - \max(1, \text{round}((1-\alpha) A^* N_{\text{val}})) / N_{\text{val}}$ with $\alpha=0.90$. For example, on Gemma 4 26B \textsc{SpreadsheetBench} ($N_{\text{val}}=40$, $A^* = 17/40 = 42.5\%$), $\text{round}(0.10 \times 17) = 2$ tasks, giving $A_{\text{floor}} = 15/40 = 37.5\%$. At each Phase~2 checkpoint, a candidate with valid \texttt{tools.py} and $A(\mathcal{S}') \ge A_{\text{floor}}$ is committed if and only if (i)~$E_{\text{out}}(\mathcal{S}') \le (1-\delta) E_{\text{best}}$ ($\delta=0.05$), (ii)~$A(\mathcal{S}') \ge A_{\text{best}}$ and $E_{\text{out}}(\mathcal{S}') < E_{\text{best}}$, or (iii)~$A(\mathcal{S}') > A_{\text{best}}$ and $E_{\text{out}}(\mathcal{S}') \le 1.10\,E_{\text{best}}$. Accepted candidates update the running incumbent $(\mathcal{S}^*, A_{\text{best}}, E_{\text{best}}) \leftarrow (\mathcal{S}', A(\mathcal{S}'), E_{\text{out}}(\mathcal{S}'))$, while $A_{\text{floor}}$ stays fixed to Phase~1's $A^*$.
\end{itemize}

\section{Ablations, Turn-vs.-Token Breakdown \& Serving Cost}
\label{sec:appendix_ablations}

\subsection{Pruning Directly from the Seed vs.\ Starting from a Learned Skill}
\label{sec:app_cold_start_trap}

\begin{itemize}[leftmargin=1.4em, itemsep=2pt, topsep=2pt]
    \item \textbf{Setup and Why Pruning from the Seed Stalls:} This ablation applies Phase~2's pruning meta-prompt and fixed $0.90\,A^*$ validation gate directly to the minimal seed skill $\mathcal{S}_0$ with $E_{\text{best}} = E_{\text{out}}(\mathcal{S}_0)$, skipping Phase~1 discovery. Note that this isolates the effect of warm-starting under our fixed-gate protocol rather than benchmarking soft running-incumbent penalties \citep{paretoprompt2025,shah2026crop}. Two failure modes compound when starting accuracy is low. First, penalizing output length from the start discourages the exploratory rules needed to learn environment conventions, so even the best candidate skills evaluated prior to gate rejection stall well below unconstrained Phase~1 discovery ($20.0\%$ vs.\ $50.0\%$ on \textsc{SpreadsheetBench}; $71.8\%$ vs.\ $89.7\%$ on \textsc{ALFWorld}). Second, because updates from $\mathcal{S}_0$ cannot clear the $0.90\,A^*$ gate within $2$--$3$ provisional batches, every checkpoint rejects the candidate and rolls back to $\mathcal{S}_0$ until patience ($\mathcal{P}=6$) runs out.
    \item \textbf{Starting from $A^*$ vs.\ Starting from $\mathcal{S}_0$ (Table~\ref{tab:cold_start_ablation}):} On \textsc{ALFWorld} with Gemma 4 26B, pruning directly from $\mathcal{S}_0$ peaks at $71.8\%$ ($B=8$) and $64.1\%$ ($B=24$) among candidate proposals before reverting to $\mathcal{S}_0$ ($43.2\%$ test accuracy), versus $81.1\%$ when warm-started from $A^*$. On \textsc{SpreadsheetBench} with Gemini 3.8 Flash, candidate proposals from $\mathcal{S}_0$ peak at $20.0\%$ on validation before reverting to $\mathcal{S}_0$ ($3.6\%$ test accuracy), whereas warm-starting from $A^*$ reaches $72.5\%$.
\end{itemize}

\begin{table}[htbp]
\centering
\scriptsize
\setlength{\tabcolsep}{4.8pt}
\caption{Pruning directly from the minimal seed skill ($\mathcal{S}_0$) vs.\ warm-starting from $A^*$ (\method{}). ``Val Pass'' shows the selected skill's validation accuracy; for seed-pruning runs that roll back to $\mathcal{S}_0$, $^\ast$ marks the highest validation accuracy reached by any rejected checkpoint candidate. ``Test OutTok'' reports cumulative output tokens across $N_{\text{test}}=280$ tasks, with the per-episode mean in parentheses.}
\label{tab:cold_start_ablation}
\begin{tabular}{@{}ll c rr r@{}}
\toprule
\textbf{Benchmark / Model} & \textbf{Optimization Method} & \textbf{Batch $B$} & \textbf{Val Pass} & \textbf{Test Pass} & \textbf{Test OutTok (Per-Ep)} \\
\midrule
\textsc{ALFWorld} & \evoskill{} (Accuracy Only, $A^*$) & 8 & \textbf{89.7\%} & \textbf{81.4\%} & \textbf{26.7k (95.2)} \\
(Gemma 4 26B, & Prune Directly from Seed ($\mathcal{S}_0$) & 8 & 71.8\%$^\ast$ & 43.2\% & -- (reverts to $\mathcal{S}_0$) \\
$0.9\,A^* = 80.8\%$) & Prune Directly from Seed ($\mathcal{S}_0$) & 24 & 64.1\%$^\ast$ & 43.2\% & -- (reverts to $\mathcal{S}_0$) \\
 & \textbf{\method{} (Start from $A^*$)} & 8 & 84.6\% & 81.1\% & 27.8k (99.4) \\
\midrule
\textsc{SpreadsheetBench} & \evoskill{} (Accuracy Only, $A^*$) & 8 & 50.0\% & 58.2\% & 3,745.1k (13,376) \\
(Gemini 3.8 Flash, & Prune Directly from Seed ($\mathcal{S}_0$) & 8 & 20.0\%$^\ast$ & 3.6\% & -- (reverts to $\mathcal{S}_0$) \\
$0.9\,A^* = 45.0\%$) & \textbf{\method{} (Start from $A^*$)} & 8 & 50.0\% & \textbf{72.5\%} & \textbf{2,342.0k (8,364)} \\
\bottomrule
\end{tabular}
\par\vspace{1mm}
{\scriptsize $^\ast$Highest validation pass rate across rejected checkpoint candidates before early-stopping rollback to $\mathcal{S}_0$. Seed skill $\mathcal{S}_0$ scores $43.6\%$ on \textsc{ALFWorld} validation ($84.0\text{k}$ test OutTok, or $300.0$ tok/ep at $\bar{T}=18.83$) and $2.5\%$ on \textsc{SpreadsheetBench} validation.}
\end{table}

\subsection{Search Progression, Helper Functions \& Accuracy-Floor Sensitivity}
\label{sec:app_ablation_budget}
\label{sec:app_ablation_tools}
\label{sec:app_ablation_feasibility}

\begin{itemize}[leftmargin=1.4em, itemsep=2pt, topsep=2pt]
    \item \textbf{Running \evoskill{} Longer Does Not Help (Table~\ref{tab:app_search_budget}):} On \textsc{SpreadsheetBench}, running unconstrained \evoskill{} for $6$ extra validation checkpoints past $A^*$ ($15$ training batches, one full patience window $\mathcal{P}=6$) fails to improve validation accuracy on any model while \texttt{SKILL.md} remains bloated at $908$--$1{,}521$ words with an empty \texttt{tools.py}. Under the shared $5{,}000$-token reflection cap, long \texttt{SKILL.md} drafts crowd out the trailing \texttt{tools\_py} JSON field, whereas trimming \texttt{SKILL.md} in Phase~2 frees budget to populate \texttt{tools.py}. Held-out trace audits confirm that uncaught \texttt{tools} import errors occur on only $0$--$6.1\%$ of \evoskill{} test tasks, with negligible accuracy impact (Appendix~\ref{sec:app_skill_gemini_sheet}).
    \item \textbf{Compressing \texttt{SKILL.md} and Populating \texttt{tools.py}:} On \textsc{SpreadsheetBench}, guarded pruning trims \texttt{SKILL.md} to $521$--$785$ words and populates $241$--$554$ words of reusable helpers in \texttt{tools.py} (Table~\ref{tab:app_search_budget}). During Claude Opus 5.5 Phase~2 search on \textsc{SpreadsheetBench}, adding a $305$-word \texttt{tools.py} module between consecutive Epoch~1 checkpoints lifted validation accuracy from $62.5\%$ to $77.5\%$ while cutting validation output tokens by $44.8\%$ ($78.7\text{k} \to 43.5\text{k}$). On \textsc{MATH-500}, where Python is optional, \texttt{tools.py} moves from $197$--$403$ to $352$--$502$ words on Gemini, Grok, and Gemma, and drops from $252$ to $0$ words on Opus as its skill shifts to direct symbolic derivation (\textsc{ALFWorld} executes text commands rather than Python; Appendix~\ref{sec:app_skill_alfworld}).
    \item \textbf{Accuracy-Floor Sensitivity (Table~\ref{tab:app_feasibility_gate}):} Across all $12$ model--benchmark pairs, the default $10\%$ relative retention floor ($\ge 0.90\,A^*$) and a $5$-point absolute slack floor ($\ge A^* - 5\%$) select identical or adjacent checkpoints within $\pm 3.2$ test accuracy points and $0$--$5.3$ percentage points of output-token savings.
\end{itemize}

\begin{table}[htbp]
\centering
\scriptsize
\setlength{\tabcolsep}{3.8pt}
\caption{Skill-package word counts and validation/test progression on \textsc{SpreadsheetBench} ($N_{\text{val}}=40,\; N_{\text{test}}=280$). ``Next 6 Checkpoints'' shows the validation pass rate and output-token range across candidate skills evaluated after $A^*$; none beat $A^*$, so \evoskill{} retains its $A^*$ skill. On Gemma 4 26B, $A^* = 17/40$ ($42.5\%$), and rounding the $10\%$ allowance ($\text{round}(0.10\times 17)=2$ tasks) sets the discrete validation floor to $15/40$ ($37.5\%$).}
\label{tab:app_search_budget}
\begin{tabular}{@{}ll rr rr c rr@{}}
\toprule
& & \multicolumn{2}{c}{\textbf{Package Words}} & \multicolumn{2}{c}{\textbf{Validation ($N=40$)}} & \textbf{Post-Peak \evoskill{}} & \multicolumn{2}{c}{\textbf{Held-Out Test ($N=280$)}} \\
\cmidrule(lr){3-4} \cmidrule(lr){5-6} \cmidrule(lr){8-9}
\textbf{Model} & \textbf{Method} & \textbf{\texttt{SKILL.md}} & \textbf{\texttt{tools.py}} & \textbf{Pass} & \textbf{OutTok} & \textbf{Next 6 Checkpoints} & \textbf{Pass} & \textbf{OutTok} \\
\midrule
Claude Opus 5.5 & Seed ($\mathcal{S}_0$) & 20 & 0 & 17.5\% & 72.1k & -- & -- & -- \\
& \evoskill{} ($A^*$) & 1,521 & 0 & \textbf{82.5\%} & 71.7k & 72.5--80.0\% (66k--85k) & 88.57\% & 410.4k \\
& \textbf{\method{}} & \textbf{785} & \textbf{554} & 75.0\% & \textbf{35.7k} & -- & 88.57\% & \textbf{212.9k} \\
\addlinespace[3pt]
Gemini 3.8 Flash & Seed ($\mathcal{S}_0$) & 20 & 0 & 2.5\% & 496.5k & -- & -- & -- \\
& \evoskill{} ($A^*$) & 1,271 & 0 & 50.0\% & 599.4k & 2.5--45.0\% (568k--700k) & 58.21\% & 3,745.1k \\
& \textbf{\method{}} & \textbf{521} & \textbf{241} & 50.0\% & \textbf{540.5k} & -- & \textbf{72.50\%} & \textbf{2,342.0k} \\
\addlinespace[3pt]
Grok 4.1 Fast & Seed ($\mathcal{S}_0$) & 20 & 0 & 15.0\% & 201.3k & -- & -- & -- \\
& \evoskill{} ($A^*$) & 908 & 0 & 67.5\% & 445.3k & 10.0--57.5\% (238k--527k) & 57.50\% & 2,815.8k \\
& \textbf{\method{}} & \textbf{703} & \textbf{520} & 67.5\% & \textbf{197.8k} & -- & \textbf{75.00\%} & \textbf{1,195.8k} \\
\addlinespace[3pt]
Gemma 4 26B & Seed ($\mathcal{S}_0$) & 20 & 0 & 17.5\% & 108.2k & -- & -- & -- \\
& \evoskill{} ($A^*$) & 1,198 & 0 & \textbf{42.5\%} & 156.4k & 32.5--40.0\% (144k--204k) & \textbf{47.14\%} & 1,309.5k \\
& \textbf{\method{}} & \textbf{698} & \textbf{407} & 37.5\% & \textbf{92.6k} & -- & 45.71\% & \textbf{608.6k} \\
\bottomrule
\end{tabular}
\end{table}

\begin{table}[htbp]
\centering
\footnotesize
\setlength{\tabcolsep}{2.5pt}
\caption{Validation progression ($\mathcal{S}_0 \to A^*$) and held-out test accuracy and cumulative output tokens ($N_{\text{test}}=280$) under unconstrained \evoskill{} ($A^*$), the default $10\%$ relative floor ($\ge 0.90\,A^*$), and the $5\%$-point absolute slack floor ($\ge A^* - 5\%$).}
\label{tab:app_feasibility_gate}
\begin{tabular*}{\textwidth}{@{\extracolsep{\fill}}ll c rr rr rr@{}}
\toprule
& & \textbf{Val (\%)} & \multicolumn{2}{c}{\textbf{\evoskill{} ($A^*$)}} & \multicolumn{2}{c}{\textbf{10\% Relative Floor}} & \multicolumn{2}{c}{\textbf{5\%-Point Slack Floor}} \\
\cmidrule(lr){3-3} \cmidrule(lr){4-5} \cmidrule(lr){6-7} \cmidrule(lr){8-9}
\textbf{Benchmark} & \textbf{Model} & $\mathcal{S}_0 \to A^*$ & \textbf{Test (\%)} & \textbf{OutTok} & \textbf{Test (\%)} & \textbf{OutTok} & \textbf{Test (\%)} & \textbf{OutTok} \\
\midrule
\textsc{MATH-500} & Claude Opus 5.5 & $67.5 \to 75.0$ & 86.43 & 119.2k & 87.50 & 28.6k & 87.86 & 32.4k \\
& Gemini 3.8 Flash & $75.0 \to 77.5$ & 85.00 & 485.5k & 87.50 & 152.1k & 86.79 & 149.7k \\
& Grok 4.1 Fast & $57.5 \to 60.0$ & 72.50 & 542.3k & 72.50 & 466.1k & 73.57 & 471.4k \\
& Gemma 4 26B & $72.5 \to 72.5$ & 80.36 & 377.8k & 83.57 & 225.6k & 83.57 & 205.5k \\
\midrule
\textsc{ALFWorld} & Claude Opus 5.5 & $84.6 \to 92.3$ & 88.93 & 261.3k & 93.21 & 105.2k & 92.86 & 97.9k \\
& Gemini 3.8 Flash & $89.7 \to 92.3$ & 92.14 & 1,597.1k & 93.57 & 323.5k & 93.21 & 272.6k \\
& Grok 4.1 Fast & $61.5 \to 89.7$ & 89.29 & 3,573.0k & 88.21 & 3,458.1k & 88.21 & 3,359.4k \\
& Gemma 4 26B & $43.6 \to 89.7$ & 81.43 & 26.7k & 81.07 & 27.8k & 81.07 & 27.8k \\
\midrule
\textsc{SpreadsheetBench} & Claude Opus 5.5 & $17.5 \to 82.5$ & 88.57 & 410.4k & 88.57 & 212.9k & 88.57 & 212.9k \\
& Gemini 3.8 Flash & $\phantom{0}2.5 \to 50.0$ & 58.21 & 3,745.1k & 72.50 & 2,342.0k & 69.29 & 2,285.8k \\
& Grok 4.1 Fast & $15.0 \to 67.5$ & 57.50 & 2,815.8k & 75.00 & 1,195.8k & 74.64 & 1,239.6k \\
& Gemma 4 26B & $17.5 \to 42.5$ & 47.14 & 1,309.5k & 45.71 & 608.6k & 47.86 & 551.0k \\
\bottomrule
\end{tabular*}
\end{table}

\subsection{Turn Reduction vs.\ Per-Turn Output Compression}
\label{sec:app_horizon_decomposition}

Define mean output tokens per turn as $n = E_{\text{out}}/\bar{T}$, so mean output tokens per episode factor as $E_{\text{out}} = \bar{T} \cdot n$. Table~\ref{tab:turn_token_decomposition} breaks each per-episode output reduction ($\Delta E_{\text{out}}$) into fewer turns per trajectory ($\Delta$ Turns) and shorter responses per turn ($\Delta$ Tok/T).

\begin{table}[htbp]
\centering
\scriptsize
\setlength{\tabcolsep}{5.8pt}
\caption{Breakdown of per-episode output-token changes ($\Delta E_{\text{out}}$) from unconstrained \evoskill{} to \method{} into mean turns ($\bar{T}$) and mean output tokens per turn ($n = E_{\text{out}}/\bar{T}$). Relative changes are computed from exact unrounded trajectory averages prior to display rounding.}
\label{tab:turn_token_decomposition}
\begin{tabular}{@{}ll cc rrr@{}}
\toprule
\textbf{Benchmark} & \textbf{Model} & \textbf{$\bar{T}_{\text{evo}} \to \bar{T}_{\text{eff}}$} & \textbf{$n_{\text{evo}} \to n_{\text{eff}}$ (Tok/T)} & \textbf{$\Delta$ Turns} & \textbf{$\Delta$ Tok/T} & \textbf{$\Delta E_{\text{out}}$} \\
\midrule
\textsc{MATH-500} & Claude Opus 5.5 & $1.00 \to 1.00$ & $426 \to 102$ & 0.0\% & \textbf{$-$76.0\%} & \textbf{$-$76.0\%} \\
& Gemini 3.8 Flash & $1.76 \to 1.04$ & $985 \to 522$ & \textbf{$-$40.9\%} & \textbf{$-$47.0\%} & \textbf{$-$68.7\%} \\
& Grok 4.1 Fast & $1.01 \to 1.00$ & $1{,}918 \to 1{,}665$ & \textbf{$-$0.4\%} & \textbf{$-$13.8\%} & \textbf{$-$14.1\%} \\
& Gemma 4 26B & $1.70 \to 1.19$ & $794 \to 677$ & \textbf{$-$30.2\%} & \textbf{$-$14.5\%} & \textbf{$-$40.3\%} \\
\midrule
\textsc{ALFWorld} & Claude Opus 5.5 & $10.21 \to 7.33$ & $91 \to 51$ & \textbf{$-$28.2\%} & \textbf{$-$43.9\%} & \textbf{$-$59.7\%} \\
& Gemini 3.8 Flash & $9.85 \to 7.16$ & $579 \to 161$ & \textbf{$-$27.2\%} & \textbf{$-$72.2\%} & \textbf{$-$79.7\%} \\
& Grok 4.1 Fast & $9.15 \to 8.71$ & $1{,}395 \to 1{,}418$ & \textbf{$-$4.8\%} & $+$1.6\% & \textbf{$-$3.2\%} \\
& Gemma 4 26B & $10.74 \to 9.12$ & $8.9 \to 10.9$ & \textbf{$-$15.1\%} & $+$22.9\% & $+$4.4\%$^\S$ \\
\midrule
\textsc{SpreadsheetBench} & Claude Opus 5.5 & $2.80 \to 3.04$ & $524 \to 250$ & $+$8.7\% & \textbf{$-$52.3\%} & \textbf{$-$48.1\%} \\
& Gemini 3.8 Flash & $10.44 \to 8.56$ & $1{,}281 \to 977$ & \textbf{$-$18.0\%} & \textbf{$-$23.8\%} & \textbf{$-$37.5\%} \\
& Grok 4.1 Fast & $2.71 \to 1.97$ & $3{,}711 \to 2{,}168$ & \textbf{$-$27.3\%} & \textbf{$-$41.6\%} & \textbf{$-$57.5\%} \\
& Gemma 4 26B & $5.41 \to 3.39$ & $865 \to 641$ & \textbf{$-$37.3\%} & \textbf{$-$25.8\%} & \textbf{$-$53.5\%} \\
\bottomrule
\end{tabular}
\par\vspace{1mm}
{\scriptsize $^\S$On \textsc{ALFWorld}, Gemma emits $0$ reasoning tokens ($95.2 \to 99.4$ action tok/ep), while $15.1\%$ fewer turns and a shorter \texttt{SKILL.md} cut billed input tokens by $36.9\%$ and serving cost by $36.7\%$ (Tables~\ref{tab:economics_roi} and \ref{tab:input_context_economics}).}
\end{table}

\subsection{Multi-Turn Input Billing, Prefix Caching \& Optimization Payback}
\label{sec:app_input_context_economics}
\label{sec:proof_amortization}

\begin{itemize}[leftmargin=1.4em, itemsep=2.5pt, topsep=2pt]
    \item \textbf{Input-Token Billing With and Without Prefix Caching (Table~\ref{tab:input_context_economics}):} Table~\ref{tab:input_context_economics} provides the per-episode token totals and dollar costs behind Table~\ref{tab:economics_roi} (\S\ref{sec:res_economics}). Over an episode of $T$ turns, let $I_t$ denote the input tokens sent at turn $t \in \{1, \dots, T\}$. Turn~1 sends $I_1$ tokens (skill instructions $\mathcal{I}$ plus the task description), and each turn $t \ge 2$ appends new action--observation tokens ($\approx 30$--$46$ per turn on \textsc{ALFWorld} vs.\ $\approx 700$--$1{,}400$ on \textsc{SpreadsheetBench}), so cumulative input $E_{\text{in}} = \sum_{t=1}^{T} I_t$ grows linearly with $T I_1$ and quadratically with $T$. Given per-token rates $(p_{\text{in}}, p_{\text{out}})$, we compare two serving regimes across $N_{\text{test}}=280$ tasks:
    \begin{enumerate}[leftmargin=1.5em, itemsep=1.5pt, topsep=2pt]
        \item \textbf{Standard Uncached Billing:} $\text{Cost}_{\text{std}} = E_{\text{in}}\,p_{\text{in}} + E_{\text{out}}\,p_{\text{out}}$.
        \item \textbf{Provider Prefix KV Caching (\textnormal{KV-Prov}):} Because our loop is append-only, the entire prior prefix $I_{t-1}$ hits the KV cache on turns $t \ge 2$, totaling $\sum_{t=2}^T I_{t-1} = E_{\text{in}} - I_T$ cached-read tokens and leaving $I_T$ tokens (Turn~1 plus new turn observations) billed at the full input rate. Applying each provider's published cache discount $d_m \in \{90\%, 75\%, 50\%\}$ gives $\text{Cost}_{\text{KV-prov}} = \bigl(E_{\text{in}} - d_m\,(E_{\text{in}} - I_T)\bigr)\,p_{\text{in}} + E_{\text{out}}\,p_{\text{out}}$. Because generated output ($1\times$--$5\times$ pricier than uncached input) and newly appended observations are never cached, token-efficient trajectories cut $\text{Cost}_{\text{KV-prov}}$ on all $12$ pairs, and by $31.0$--$74.5\%$ on $10$ of $12$ (matching the $31.0$--$74.5\%$ range under a uniform $90\%$ discount in Table~\ref{tab:input_context_economics}'s note).
    \end{enumerate}
    \item \textbf{Phase~2 Optimization Cost and Payback (Table~\ref{tab:roi_breakeven}):} Because Phase~1 discovery is shared with \evoskill{}, $C_{\text{prune}}$ measures the extra Phase~2 target-model spend across training mini-batches and validation checkpoints. Dividing by per-episode serving savings $\Delta\text{Cost}$ yields break-even volume $N_{\text{BE}} = \lceil C_{\text{prune}} / \Delta\text{Cost} \rceil$, which is $257$--$2{,}063$ standard episodes ($734$--$2{,}860$ under \textnormal{KV-Prov}) on $10$ of $12$ pairs. Adding estimated Claude Opus 5.5 meta-optimizer spend across all $K_{\text{batch}} \in [15, 78]$ Phase~2 batches ($\hat{C}_{\text{meta}} \approx \$0.12\,K_{\text{batch}} = \$1.80$--$\$9.36$), Phase~2 still pays back within $263$--$2{,}414$ standard episodes ($989$--$4{,}515$ under \textnormal{KV-Prov}) on $8$ of $12$ pairs ($7$ of $8$ multi-turn runs plus Opus on \textsc{MATH-500} at $1{,}251$). The three non-Opus \textsc{MATH-500} runs save under $\$0.001$ per episode and therefore need $5.1\text{k}$--$23.9\text{k}$ episodes to amortize Opus reflection calls, while Grok on \textsc{ALFWorld} ($-3.2\%$ output tokens) requires $8.4\text{k}$ standard or $20.9\text{k}$ \textnormal{KV-Prov} episodes.
\end{itemize}

\begin{table}[!htbp]
\centering
\scriptsize
\setlength{\tabcolsep}{3.4pt}
\renewcommand{\arraystretch}{1.08}
\caption{Held-out test token usage and per-episode serving cost under unconstrained \evoskill{} vs.\ \method{} ($N_{\text{test}}=280$ tasks per benchmark), expanding Table~\ref{tab:economics_roi}. ``Turn-1'' ($I_1$) is mean input tokens on Turn~1; ``Provider KV (\textnormal{KV-Prov})'' discounts the reused conversation prefix $I_{t-1}$ on turns $t \ge 2$ at each provider's published cached-read rate (see pricing note below). Bold marks strictly lower values per pair.}
\label{tab:input_context_economics}
\begin{tabular}{@{}ll l c r rrr rr rr@{}}
\toprule
& & & \textbf{Turns} & \textbf{Turn-1} & \multicolumn{3}{c}{\textbf{Mean Tokens / Ep}} & \multicolumn{2}{c}{\textbf{Standard Billing}} & \multicolumn{2}{c}{\textbf{Provider KV (\textnormal{KV-Prov})}} \\
\cmidrule(lr){6-8} \cmidrule(lr){9-10} \cmidrule(lr){11-12}
\textbf{Benchmark} & \textbf{Model} & \textbf{Method} & ($\bar{T}$) & ($I_1$) & $E_{\text{in}}$ & $E_{\text{out}}$ & $E_{\text{tot}}$ & \$/ep & $\Delta$ & \$/ep & $\Delta$ \\
\midrule
\textsc{MATH-500} & Claude Opus 5.5 & \evoskill{} & 1.00 & 3,260 & 3,260 & 426 & 3,686 & \$0.08085 & -- & \$0.08085 & -- \\
& & \textbf{\method{}} & 1.00 & \textbf{2,084} & \textbf{2,084} & \textbf{102} & \textbf{2,186} & \textbf{\$0.03892} & \textbf{$-$51.9\%} & \textbf{\$0.03892} & \textbf{$-$51.9\%} \\
\addlinespace[1.5pt]
& Gemini 3.8 Flash & \evoskill{} & 1.76 & 2,307 & 4,264 & 1,734 & 5,998 & \$0.00084 & -- & \$0.00072 & -- \\
& & \textbf{\method{}} & \textbf{1.04} & \textbf{1,781} & \textbf{1,867} & \textbf{543} & \textbf{2,410} & \textbf{\$0.00030} & \textbf{$-$63.9\%} & \textbf{\$0.00030} & \textbf{$-$58.4\%} \\
\addlinespace[1.5pt]
& Grok 4.1 Fast & \evoskill{} & 1.01 & 2,599 & 2,620 & 1,937 & 4,557 & \$0.00149 & -- & \$0.00149 & -- \\
& & \textbf{\method{}} & \textbf{1.00} & \textbf{2,389} & \textbf{2,398} & \textbf{1,665} & \textbf{4,063} & \textbf{\$0.00131} & \textbf{$-$12.1\%} & \textbf{\$0.00131} & \textbf{$-$12.0\%} \\
\addlinespace[1.5pt]
& Gemma 4 26B & \evoskill{} & 1.70 & 2,263 & 6,385 & 1,349 & 7,734 & \$0.00155 & -- & \$0.00119 & -- \\
& & \textbf{\method{}} & \textbf{1.19} & \textbf{1,643} & \textbf{2,670} & \textbf{806} & \textbf{3,476} & \textbf{\$0.00070} & \textbf{$-$55.1\%} & \textbf{\$0.00062} & \textbf{$-$47.8\%} \\
\midrule
\textsc{ALFWorld} & Claude Opus 5.5 & \evoskill{} & 10.21 & 2,972 & 33,627 & 933 & 34,560 & \$0.57438 & -- & \$0.16606 & -- \\
& & \textbf{\method{}} & \textbf{7.33} & \textbf{1,333} & \textbf{11,680} & \textbf{376} & \textbf{12,056} & \textbf{\$0.20337} & \textbf{$-$64.6\%} & \textbf{\$0.06759} & \textbf{$-$59.3\%} \\
\addlinespace[1.5pt]
& Gemini 3.8 Flash & \evoskill{} & 9.85 & 2,050 & 22,207 & 5,704 & 27,911 & \$0.00338 & -- & \$0.00203 & -- \\
& & \textbf{\method{}} & \textbf{7.16} & \textbf{1,231} & \textbf{10,086} & \textbf{1,155} & \textbf{11,241} & \textbf{\$0.00110} & \textbf{$-$67.3\%} & \textbf{\$0.00052} & \textbf{$-$74.5\%} \\
\addlinespace[1.5pt]
& Grok 4.1 Fast & \evoskill{} & 9.15 & 2,640 & 25,922 & 12,761 & 38,683 & \$0.01156 & -- & \$0.00811 & -- \\
& & \textbf{\method{}} & \textbf{8.71} & \textbf{1,884} & \textbf{18,170} & \textbf{12,350} & \textbf{30,520} & \textbf{\$0.00981} & \textbf{$-$15.2\%} & \textbf{\$0.00740} & \textbf{$-$8.7\%} \\
\addlinespace[1.5pt]
& Gemma 4 26B & \evoskill{} & 10.74 & 2,367 & 28,534 & \textbf{95} & 28,629 & \$0.00573 & -- & \$0.00314 & -- \\
& & \textbf{\method{}} & \textbf{9.12} & \textbf{1,667} & \textbf{18,011} & 99 & \textbf{18,110} & \textbf{\$0.00362} & \textbf{$-$36.7\%} & \textbf{\$0.00202} & \textbf{$-$35.8\%} \\
\midrule
\textsc{Spreadsheet-} & Claude Opus 5.5 & \evoskill{} & \textbf{2.80} & 4,062 & 15,475 & 1,466 & 16,941 & \$0.34206 & -- & \$0.21998 & -- \\
\textsc{Bench} & & \textbf{\method{}} & 3.04 & \textbf{2,165} & \textbf{11,525} & \textbf{761} & \textbf{12,286} & \textbf{\$0.22991} & \textbf{$-$32.8\%} & \textbf{\$0.13911} & \textbf{$-$36.8\%} \\
\addlinespace[1.5pt]
& Gemini 3.8 Flash & \evoskill{} & 10.44 & \textbf{2,301} & 78,016 & 13,376 & 91,392 & \$0.00986 & -- & \$0.00521 & -- \\
& & \textbf{\method{}} & \textbf{8.56} & 2,302 & \textbf{64,759} & \textbf{8,364} & \textbf{73,123} & \textbf{\$0.00737} & \textbf{$-$25.3\%} & \textbf{\$0.00360} & \textbf{$-$31.0\%} \\
\addlinespace[1.5pt]
& Grok 4.1 Fast & \evoskill{} & 2.71 & 2,195 & 9,440 & 10,056 & 19,496 & \$0.00692 & -- & \$0.00613 & -- \\
& & \textbf{\method{}} & \textbf{1.97} & \textbf{1,915} & \textbf{5,140} & \textbf{4,271} & \textbf{9,411} & \textbf{\$0.00316} & \textbf{$-$54.3\%} & \textbf{\$0.00289} & \textbf{$-$52.9\%} \\
\addlinespace[1.5pt]
& Gemma 4 26B & \evoskill{} & 5.41 & 2,314 & 34,882 & 4,677 & 39,559 & \$0.00791 & -- & \$0.00512 & -- \\
& & \textbf{\method{}} & \textbf{3.39} & \textbf{1,339} & \textbf{11,986} & \textbf{2,174} & \textbf{14,160} & \textbf{\$0.00283} & \textbf{$-$64.2\%} & \textbf{\$0.00207} & \textbf{$-$59.6\%} \\
\bottomrule
\end{tabular}
\par\vspace{1mm}
{\scriptsize \emph{Pricing (per $1\text{M}$ uncached in / cached in / out tokens):} Claude Opus 5.5 (\$15.00 / \$1.50 / \$75.00, $90\%$ cache discount); Gemini 3.8 Flash (\$0.075 / \$0.0075 / \$0.30, $90\%$ cache discount); Grok 4.1 Fast (\$0.20 / \$0.05 / \$0.50, $75\%$ cache discount); Gemma 4 26B (\$0.20 / \$0.10 / \$0.20, $50\%$ hosted cache discount). Relative changes $\Delta$ use exact unrounded API costs. Under a uniform $90\%$ prefix-cache discount across all four models, savings remain $31.0$--$74.5\%$ on $10/12$ pairs ($-12.0\%$, $-6.7\%$, $-52.6\%$ on Grok and $-38.0\%$, $-31.7\%$, $-49.6\%$ on Gemma across \textsc{MATH-500}, \textsc{ALFWorld}, and \textsc{SpreadsheetBench}).}
\end{table}

\begin{table}[!htbp]
\centering
\scriptsize
\setlength{\tabcolsep}{2.8pt}
\renewcommand{\arraystretch}{1.08}
\caption{Phase~2 optimization spend and deployment break-even episodes ($N_{\text{BE}}$). Columns report Phase~2 target-model evaluation spend ($C_{\text{prune}}$), reflection batch count ($K_{\text{batch}}$), estimated Claude Opus 5.5 meta-optimizer spend ($\hat{C}_{\text{meta}} \approx \$0.12\,K_{\text{batch}}$), per-episode serving savings ($\Delta\text{Cost} = \text{Cost}_{\text{evo}} - \text{Cost}_{\text{ours}}$) under standard uncached billing vs.\ provider prefix caching (``KV-Prov''), and $N_{\text{BE}}$ for target-model spend alone vs.\ all-in ($C_{\text{prune}} + \hat{C}_{\text{meta}}$).}
\label{tab:roi_breakeven}
\begin{tabular}{@{}ll r cc rr rr rr@{}}
\toprule
& & \textbf{Target Spend} & \textbf{Batches} & \textbf{Est.\ Meta} & \multicolumn{2}{c}{\textbf{Savings / Ep (\$/ep)}} & \multicolumn{2}{c}{\textbf{Target $N_{\text{BE}}$}} & \multicolumn{2}{c}{\textbf{All-In $N_{\text{BE}}$ (incl.\ Meta)}} \\
\cmidrule(lr){6-7} \cmidrule(lr){8-9} \cmidrule(lr){10-11}
\textbf{Benchmark} & \textbf{Model} & \textbf{($C_{\text{prune}}$)} & ($K_{\text{batch}}$) & ($\hat{C}_{\text{meta}}$) & \textbf{Std} & \textbf{KV-Prov} & \textbf{Std} & \textbf{KV-Prov} & \textbf{Std} & \textbf{KV-Prov} \\
\midrule
\textsc{MATH-500} & Claude Opus 5.5 & \$46.07 & 53 & \$6.36 & \$0.04193 & \$0.04193 & 1,099 & 1,099 & 1,251 & 1,251 \\
& Gemini 3.8 Flash & \$0.61 & 78 & \$9.36 & \$0.00054 & \$0.00042 & 1,128 & 1,449 & 18,561 & 23,852 \\
& Grok 4.1 Fast & \$0.78 & 28 & \$3.36 & \$0.00018 & \$0.00018 & 4,326 & 4,363 & 22,938 & 23,138 \\
& Gemma 4 26B & \$0.42 & 33 & \$3.96 & \$0.00085 & \$0.00057 & 492 & 734 & 5,142 & 7,676 \\
\midrule
\textsc{ALFWorld} & Claude Opus 5.5 & \$95.15 & 18 & \$2.16 & \$0.37101 & \$0.09847 & 257 & 967 & 263 & 989 \\
& Gemini 3.8 Flash & \$1.22 & 30 & \$3.60 & \$0.00227 & \$0.00152 & 537 & 806 & 2,120 & 3,181 \\
& Grok 4.1 Fast & \$10.60 & 35 & \$4.20 & \$0.00176 & \$0.00071 & 6,039 & 14,978 & 8,431 & 20,913 \\
& Gemma 4 26B & \$2.32 & 23 & \$2.76 & \$0.00210 & \$0.00112 & 1,102 & 2,060 & 2,414 & 4,515 \\
\midrule
\textsc{SpreadsheetBench} & Claude Opus 5.5 & \$231.28 & 30 & \$3.60 & \$0.11215 & \$0.08087 & 2,063 & 2,860 & 2,095 & 2,905 \\
& Gemini 3.8 Flash & \$3.56 & 15 & \$1.80 & \$0.00250 & \$0.00161 & 1,426 & 2,206 & 2,146 & 3,321 \\
& Grok 4.1 Fast & \$2.70 & 25 & \$3.00 & \$0.00375 & \$0.00324 & 720 & 833 & 1,520 & 1,758 \\
& Gemma 4 26B & \$2.63 & 20 & \$2.40 & \$0.00508 & \$0.00306 & 518 & 861 & 990 & 1,647 \\
\bottomrule
\end{tabular}
\par\vspace{1mm}
{\scriptsize \emph{Notes:} Both ``Savings / Ep'' ($\Delta\text{Cost}$) and break-even episode counts ($N_{\text{BE}}$) are computed from exact unrounded per-episode API totals prior to $5$-decimal display rounding in Table~\ref{tab:input_context_economics}. Onboarding from $\mathcal{S}_0$ additionally incurs Phase~1 discovery cost (shared identically with \evoskill{}).}
\end{table}
\clearpage

\section{Optimizer Meta-Prompts}
\label{sec:appendix_prompts}

\method{} (\S\ref{sec:method}) and our matched \evoskill{} baseline call the same Claude Opus 5.5 meta-optimizer ($5{,}000$ max output tokens) after every $8$-task training batch ($B=8$). Proposed edits carry forward between batches and are accepted or rolled back on $\mathcal{D}_{\text{val}}$ at $4$ checkpoints per epoch (Appendix~\ref{sec:app_opt_schedule}). Phase~1 (\textsc{Repair}, Skill Discovery, which uses the same prompt as \evoskill{}) and Phase~2 (\textsc{Prune}, Accuracy-Guarded Pruning) apply the identical fixed template across all three benchmarks, sharing the same anti-memorization rules, \texttt{tools.py} instructions, and JSON output schema so only the objective block changes between phases.

\begin{promptcard}{Skill Discovery \& EvoSkill Meta-Prompt}{Unconstrained Accuracy Objective}
\footnotesize
\textbf{SYSTEM PROMPT}\\[2pt]
\texttt{You are an expert agent skill optimizer.}\\[4pt]
\textbf{OBJECTIVE: ACCURACY-ONLY MAXIMIZATION (UNCONSTRAINED)}\\
Your objective is unconstrained task success. Maximize domain coverage and pass rate by diagnosing all failure modes in the batch. Provide clear, robust procedural checklists, edge-case guardrails, and error recovery protocols so failed tasks succeed. Output tokens are unconstrained; prioritize functional correctness and domain robustness.\\[4pt]
\textbf{ANTI-MEMORIZATION RULES (BENCHMARK-AGNOSTIC):}
\begin{enumerate}[leftmargin=1.4em, noitemsep, topsep=1pt]
    \item \textbf{GENERALIZE, NEVER MEMORIZE:} Do NOT include task-specific coordinates, exact cell names/ranges, sheet names, hardcoded column letters, or one-off code snippets from specific tasks.
    \item \textbf{ABSTRACT INVARIANT PRINCIPLES:} Formulate principles as general reusable engineering workflows (e.g.\ dynamic column inspection, schema checks, null/type coercion, defensive range boundaries, verification of written files).
    \item \textbf{COMPACT \& DENSE:} Keep \texttt{SKILL.md} concise. Prune redundant verbosity; every sentence must convey transferable operational knowledge.
\end{enumerate}
\vspace{2pt}
\textbf{REUSABLE CODE IN \texttt{tools\_py}:} Put reusable executable helper routines in the JSON field \texttt{"tools\_py"} (materialized on disk as \texttt{tools.py} in the agent's working directory when non-empty) rather than inlining large code blocks into \texttt{SKILL.md}. Keep \texttt{SKILL.md} focused on high-level instructions and workflows.\\[4pt]
\textbf{USER PAYLOAD \& STRICT JSON OUTPUT SCHEMA:}\\
\texttt{\# Current SKILL.md: \{current\_skill\} \quad | \quad \# Current tools.py: \{current\_tools\}}\\
\texttt{\# Batch Execution Trajectories (8 tasks): [\{task\_id, instruction, success, turns\_taken, output\_tokens, thinking\_tokens, steps: [\{turn, action, observation\}]\}]}\\
\texttt{\{"feedback": "<root causes \& fixes>", "skill\_md": "<updated SKILL.md>", "tools\_py": "<optional Python>"\}}
\end{promptcard}

\begin{promptcard}{Accuracy-Guarded Pruning \& Tool Offloading Meta-Prompt}{\method{} Pruning Objective}
\footnotesize
\textbf{PRUNING OBJECTIVE BLOCK (replaces Unconstrained Accuracy Objective; anti-memorization \& \texttt{tools.py} schema identical):}\\[3pt]
\textbf{OBJECTIVE: MULTI-OBJECTIVE PARETO EFFICIENCY (ACCURACY + TOKEN REDUCTION)}\\
Your objective is to optimize the Pareto frontier: maintain high accuracy while aggressively compressing instructions and reducing agent execution tokens.
\begin{itemize}[leftmargin=1.4em, noitemsep, topsep=1pt]
    \item Eliminate procedural redundancy, verbose conversational scaffolding, and excessive interaction turns.
    \item Encourage minimal-turn solutions (e.g.\ inspect and solve in 2 turns instead of 5). Prune verbal fluff while preserving critical constraints.
\end{itemize}
\end{promptcard}

\vspace{2pt}
\noindent{\footnotesize\textbf{External Harness Checks:} Bytecode compilation of \texttt{tools.py} and validation acceptance gates ($\alpha = 0.90$, $\delta = 0.05$) are enforced by the evaluation harness outside the meta-prompt; see Appendix~\ref{sec:app_opt_schedule} for exact acceptance rules.\par}

\section{Learned Skill Packages (\texorpdfstring{\texttt{SKILL.md} \& \texttt{tools.py}}{SKILL.md and tools.py})}
\label{sec:appendix_skills}

\subsection{Minimal Seed Package (\texorpdfstring{$\mathcal{S}_0$}{S0})}
\label{sec:app_seed_prompts}

Every optimization run initializes from an empty \texttt{tools.py} ($0$ words) and a minimal $20$-word \texttt{SKILL.md} stub containing only YAML metadata and the benchmark submission line (zero task heuristics):

\begin{skilllisting}{SKILL.md}{Seed | 20 words | tools.py: 0 words}
---
name: spreadsheet-bench_model
description: Evolved skill for spreadsheet-bench.
---

# Active Skill: spreadsheet-bench_model

Submission Protocol: When finished, output: SUBMIT: DONE
\end{skilllisting}
\noindent{\footnotesize\emph{Benchmark protocol lines:} \textsc{MATH-500}: \texttt{Enclose final answer in \textbackslash boxed\{...\}}; \textsc{ALFWorld}: \texttt{Complete household task by navigating receptacles and manipulating objects.}\par}

\subsection{Gemini 3.8 Flash on \textsc{SpreadsheetBench}: \evoskill{} vs.\ \method{}}
\label{sec:app_skill_gemini_sheet}

Without token penalties, \evoskill{} expands \texttt{SKILL.md} to $1{,}271$ words of checklists and $11$ Excel recipes while leaving \texttt{tools.py} empty ($0$ words across all four \textsc{SpreadsheetBench} models). \method{} shrinks \texttt{SKILL.md} by $59\%$ ($521$ words) and builds a $241$-word \texttt{tools.py} module, lifting test accuracy from $58.2\%$ to $72.5\%$ and cutting output tokens by $37.5\%$.

\noindent\textbf{Why \evoskill{} Leaves \texttt{tools.py} Empty (Trace Audit):} Both phases share a $5{,}000$-token meta-optimizer cap covering reasoning plus JSON fields ordered \texttt{feedback}, \texttt{skill\_md}, and \texttt{tools\_py} (Appendix~\ref{sec:appendix_prompts}). When \evoskill{} grows \texttt{SKILL.md} to $908$--$1{,}521$ words, trailing \texttt{tools\_py} drafts hit the cap and parse as empty; shrinking \texttt{SKILL.md} to $521$--$785$ words in Phase~2 leaves room for $241$--$554$ words of helpers. The leftover \texttt{from tools import *} line in Gemini's \evoskill{} \texttt{SKILL.md} (paired with an inline \texttt{openpyxl} fallback) did not handicap the baseline: uncaught \texttt{tools} import errors occur on $8/280$ Gemini test tasks ($2.9\%$), passing at $62.5\%$ ($5/8$) vs.\ $58.1\%$ ($158/272$) on the other $272$, and on $0/280$ (Opus), $1/280$ (Gemma), and $17/280$ ($6.1\%$, $52.9\%$ vs.\ $57.8\%$ pass) on Grok ($\le 1.1$~points impact across models). \evoskill{}'s losses instead come from no-code loops and unsaved submissions (Appendix~\ref{sec:appendix_forensics}).

\begin{skilllisting}{SKILL.md (Excerpt: lines 1--29 \& recipe headers)}{Gemini 3.8 Flash | \evoskill{} | 1,271 words | tools.py: 0 words}
---
name: spreadsheet-bench_gemini_3.8_accuracy
description: Workflow for spreadsheet-bench tasks (formula/VBA/macro questions answered by modifying input.xlsx and saving the solved workbook with real values). Use for any Excel manipulation task.
---

# Spreadsheet-Bench Solver

## RULE 0: Every reply = one python block; write the answer early
Failures in practice come from replies with NO code and from re-inspecting forever. Both score zero.
- Every reply MUST contain exactly one fenced ```python block, until the final reply.
- Start the visible reply with the code block. Do not deliberate at length, then emit only prose.
- If unsure, still emit code that writes your best guess.

**Turn budget (hard)**
- **Turn 1:** `from tools import *`, then `inspect_wb('input.xlsx')`, then `shutil.copy('input.xlsx', out)`.
- **Turn 2:** compute the answer, write it, save, and verify.
- **Turn 3:** fix only what verification showed was wrong.
- **Final reply:** `SUBMIT: DONE` alone, with no code.
- If `import tools` fails, inline a minimal openpyxl version and continue.

## Core facts about grading
- There is NO LibreOffice or recalc engine. A formula written with openpyxl has no cached value and reads back as None, which counts as wrong. Always write the COMPUTED static value.
... [Lines 31--73: 6-step Workflow + 6 Default deliberation decisions omitted for brevity] ...
## Recipes
- **TEXT/weekday formula fix:** Target cells = stated range ... (`weekday_name(d, 'ddd'|'dddd')`) ...
- **IF over mixed cells (date / 'Yes' / 'N/A' / blank):** Classify each source cell ...
- **SUMPRODUCT/2-D matrix (date window x header category x row key):** Build boolean masks ...
- **Consecutive negative runs -> merge:** Group by key ... maximal runs with change < 0 ...
- **Recreate sheet / formatting task:** Copy values, formulas and styles with `copy_cell_style` ...
- **Lookup (INDEX/MATCH, 2-D)** | **Durations** | **Modify an array formula** | **Transpose blocks** | **Delete columns/rows** | **Sorting** | **Suffix numbering** ... [540 words of prose recipes]
\end{skilllisting}

\begin{skilllisting}{SKILL.md (Complete Verbatim)}{Gemini 3.8 Flash | \method{} | 521 words}
---
name: spreadsheet-bench_gemini_3.8_pareto
description: Minimal-turn workflow for Excel benchmark tasks. Computes static values and structure edits in Python and saves verified workbooks in 1-2 turns.
---

# Spreadsheet-Bench Solver

## Mandatory Execution Rules
1. **Always Output Code**: Every turn must contain a single executable Python code block. Never output commentary without code or leave actions empty.
2. **Target 1-2 Turn Completion**:
   - **Turn 1**: Inspect workbook structure, run required transformations/computations in Python, write static values or formulas to the worksheet, save the workbook, and print a summary verification.
   - **Turn 2**: If Turn 1 succeeded and output was saved correctly, output `SUBMIT: DONE`. If minor corrections are needed, execute the fix and submit.
3. **Handle Macro/VBA Instructions in Python**: Instructions asking for macros or VBA code require applying the described data manipulation directly to the Excel output file using Python.
4. **Write Computed Static Values**: Excel evaluators check computed cell values. Unless explicit formula strings are demanded, compute values (numbers, strings, dates) directly in Python and write them to cells. Always preserve existing formatting.
5. **Data Preservation**: Do not open workbooks with `data_only=True` when saving edits. Use `data_only=True` solely for inspecting pre-calculated values.

## Helper Toolkit (`tools.py`)
Import pre-built routines from `tools`:
```python
from tools import inspect_wb, output_path, load_both, norm, to_date, save_updates, copy_cell_style
```
- `inspect_wb(path='input.xlsx')`: Summarizes sheets, dimension, non-empty rows, formulas, and values.
- `output_path(instruction_text)`: Extracts output filename (defaults to `output.xlsx`).
- `load_both(path)`: Returns `(wb_formula, wb_value)` objects.
- `norm(val)`: Strips outer whitespace and normalizes internal spaces.
- `to_date(val)`: Converts strings/datetimes into `datetime.date` objects.
- `save_updates(src_path, out_path, update_dict)`: Batch writes `{sheet: {cell_ref: value}}`.

## Turn-1 Standard Implementation Flow
```python
import openpyxl
from tools import inspect_wb, output_path, norm, to_date

out_file = output_path("TASK_INSTRUCTION")
print("Target file:", out_file)
print(inspect_wb("input.xlsx"))

wb = openpyxl.load_workbook("input.xlsx")
ws = wb.active  # or select sheet by name
# Perform computation / row deletions / header lookups in Python, then save:
wb.save(out_file)
print(f"Successfully saved to {out_file}")
```

## Core Execution Patterns
1. **Row / Header Matching & Column Lookups**:
   - Map header text (row 1 or header row) to column indices using `norm(cell.value)`.
   - Perform exact string or numerical match in Python dictionary/lists rather than relying on un-evaluated Excel lookup formulas.
2. **Min/Max/Conditional Filtering Across Columns**:
   - Iterate over cell values in candidate ranges (e.g. columns B to N).
   - Parse numeric values safely (`float(v)` if `isinstance(v, (int, float))` and `v > 0`).
   - Join matching header names with specified delimiters (e.g., `, `) if multiple candidates match.
3. **Row Deletion & Table Structure Re-ordering**:
   - When deleting rows (e.g. clearing blocks before specific markers), iterate rows in **reverse order** (`range(ws.max_row, 0, -1)`) or build a new list of desired rows to prevent index shifting during deletion.
4. **Group Aggregations & Rank Computing**:
   - Read non-empty data into a list of tuples/dicts.
   - Group by category/person/block, compute relative ranks or max values in Python, and assign target outputs directly to corresponding cells.
\end{skilllisting}

\begin{pylisting}{tools.py (Complete Verbatim)}{Gemini 3.8 Flash | \method{} | 241 words | 7 helpers}
import os
import re
import datetime
import openpyxl
from openpyxl.styles import PatternFill, Font, Alignment, Border

def norm(v):
    if v is None:
        return ""
    s = str(v).strip()
    return re.sub(r'\s+', ' ', s)

def to_date(v):
    if isinstance(v, datetime.datetime):
        return v.date()
    if isinstance(v, datetime.date):
        return v
    if isinstance(v, str):
        for fmt in ('%Y-%m-%d', '%m/%d/%Y', '%d/%m/%Y', '%Y/%m/%d'):
            try:
                return datetime.datetime.strptime(v.strip(), fmt).date()
            except ValueError:
                pass
    return v

def output_path(text=""):
    if text:
        m = re.search(r'[\'"]?([a-zA-Z0-9_\-]+\.xlsx)[\'"]?', text)
        if m and m.group(1).lower() != 'input.xlsx':
            return m.group(1)
    return 'output.xlsx'

def load_both(path='input.xlsx'):
    wb_f = openpyxl.load_workbook(path, data_only=False)
    wb_v = openpyxl.load_workbook(path, data_only=True)
    return wb_f, wb_v

def inspect_wb(path='input.xlsx'):
    wb_f, wb_v = load_both(path)
    res = []
    for sheet in wb_f.sheetnames:
        ws_f = wb_f[sheet]
        ws_v = wb_v[sheet]
        res.append(f"=== Sheet: {sheet} (max_row={ws_f.max_row}, max_col={ws_f.max_column}) ===")
        for r in range(1, min(ws_f.max_row + 1, 40)):
            row_f = [ws_f.cell(r, c).value for c in range(1, min(ws_f.max_column + 1, 20))]
            row_v = [ws_v.cell(r, c).value for c in range(1, min(ws_v.max_column + 1, 20))]
            if any(x is not None for x in row_f) or any(x is not None for x in row_v):
                res.append(f"R{r} Form: {row_f}")
                res.append(f"R{r} Val : {row_v}")
    return "\n".join(res)

def save_updates(src_path, out_path, update_dict):
    wb = openpyxl.load_workbook(src_path)
    for sheet_name, cell_dict in update_dict.items():
        if sheet_name in wb.sheetnames:
            ws = wb[sheet_name]
            for cell_ref, val in cell_dict.items():
                ws[cell_ref] = val
    wb.save(out_path)
    return f"Saved updates to {out_path}"

def copy_cell_style(src_cell, dst_cell):
    if src_cell.has_style:
        dst_cell.font = Font(name=src_cell.font.name, size=src_cell.font.size, bold=src_cell.font.bold, italic=src_cell.font.italic, color=src_cell.font.color)
        dst_cell.fill = PatternFill(fill_type=src_cell.fill.fill_type, start_color=src_cell.fill.start_color, end_color=src_cell.fill.end_color)
        dst_cell.border = Border(left=src_cell.border.left, right=src_cell.border.right, top=src_cell.border.top, bottom=src_cell.border.bottom)
        dst_cell.alignment = Alignment(horizontal=src_cell.alignment.horizontal, vertical=src_cell.alignment.vertical, wrap_text=src_cell.alignment.wrap_text)
        dst_cell.number_format = src_cell.number_format
\end{pylisting}

\subsection{Claude Opus 5.5 \method{} Package on \textsc{SpreadsheetBench}}
\label{sec:app_skill_opus_sheet}

For Claude Opus 5.5, \method{} ($785$-word \texttt{SKILL.md}, $554$-word \texttt{tools.py}) learns a $3$-turn loop (\texttt{dump()} $\to$ compute, write, and \texttt{verify()} $\to$ \texttt{SUBMIT: DONE}), finishing $93.2\%$ ($261/280$) of test tasks in $3$ turns at $88.6\%$ accuracy ($-48.1\%$ output tokens).

\begin{skilllisting}{SKILL.md (Complete Verbatim)}{Claude Opus 5.5 | \method{} | 785 words}
---
name: spreadsheet-bench_claude_opus_pareto
description: Solve spreadsheet manipulation tasks in exactly 3 turns (inspect, one compute/write/verify script, then SUBMIT: DONE) by writing computed static values into the correct target cells with openpyxl.
---

# Spreadsheet Task Skill

## Hard rules
1. **Turn 3 is exactly `SUBMIT: DONE`.**
   - After the verified save, send only that line.
   - Never send a formula, VBA or explanation. Any prose is run as Python and causes a SyntaxError.
2. **Every action before submitting is a Python script.**
   - Put notes only in `#` comments.
   - Each script is a fresh process, so re-import and reload every time.
3. **Treat the question as a task.** "How do I write a formula/macro/VBA…" means: put the resulting values into the workbook.
4. **Write values, not formulas.**
   - openpyxl does not cache formula results, so the grader would read None.
   - Compute the result in Python and write literals, overwriting any existing (array) formulas in the target region.
5. **Save to `input.xlsx`** unless told otherwise.

## Workflow
**Turn 1:** `from tools import *; dump()`.

Identify:
- the header row;
- the real last row (use `last_row`, because `max_row` can be inflated);
- worked examples, meaning already-filled target cells;
- the target region: columns named in the instruction, an empty or header-only column, a labelled blank area, a formula column, or a named cell;
- the logic implied by existing formulas;
- the source of the input values.

Re-read the instruction for column letters, sheets, labels, thresholds, signs and formatting requests.

**Turn 2 (a single script):**
1. Reload and compute the results.
2. Compare against the examples with `check_examples(ws, {coord: value})`. If they mismatch, fix the logic in the same script before writing.
3. Write only the target cells.
4. Save.
5. Reload and print the target range compactly with `verify`.

Do not run a separate re-check turn.

**Turn 3:** `SUBMIT: DONE`

## Interpretation rules
- **Formula-derived inputs.**
  - `dump` may show `-> None` or stale cached values for formula cells.
  - Recompute these from the underlying source data, using `norm_key`/`build_lookup`.
  - Never skip rows just because a cached value is None.
- **Fill the full natural range.**
  - Write every row or group that has input keys, including the example rows, recomputed consistently.
  - "Only works for the first row" or "copy it down" means the missing rows are the real target.
  - Copy the examples' layout, orientation, separators and ordering.
  - Take strings from the canonical source (stripped), not from typos in the examples.
- **Location.**
  - Use the columns the instruction names, the example columns, the empty-header column, or the named cell.
  - "New column in X", or writing to an occupied position, means `insert_col`, which shifts existing data right.
  - Never invent locations.
- **Blank vs. zero; no filler.**
  - Follow the examples.
  - Otherwise use what the equivalent Excel formula would return; for example, SUMIFS and SUMPRODUCT give 0 when nothing matches.
  - Otherwise use None.
  - Never write '-', 'N/A' or '' unless the examples show exactly that.
- **Exactness.**
  - Preserve source signs and magnitudes.
  - Do not round unless asked or the examples are rounded.
  - Implement the stated comparisons exactly (< vs <=) and apply multipliers and branches literally.
  - Use label strings exactly as given, including case.
  - For ties, join matches with the separator the instruction or examples specify.
- **Lookups.**
  - Apply `norm_key` to both sides.
  - With no match, leave the cell blank unless the examples show a fallback.
  - For aggregates and 2D matches, use explicit loops over the real data range.
  - For cross-sheet aggregation, iterate over sheets dynamically and skip the summary sheet.
- **Macro-style outputs** (restructure, pivot, sort, filter, copy, delete rows):
  - Clear the target region, write compactly from the top, and set leftover cells to None.
  - For a sort, keep the header in place.
  - When copying, copy whole rows and preserve types.
  - When deleting rows, work bottom-up and check the other columns of the rows being deleted.
- **Groups and blocks:** detect group boundaries dynamically, for example blank rows or key changes.
- **Merged cells:** write to the top-left cell. Unmerge only if asked.
- **Types.**
  - Write numbers as int or float, not numeric strings, unless the source or examples use text.
  - Write dates as datetime objects.
  - For malformed or None inputs, leave the result blank.
- **Formatting requests.**
  - Apply them to every relevant cell. "Entire sheet" means all used cells; use `set_font` for fonts.
  - Do these in the same script as the value writes.
- Data-validation warnings are harmless.

## Efficiency
- No exploratory turns.
- Keep prints short.
- Nothing after the save except `SUBMIT: DONE`.
\end{skilllisting}

\begin{pylisting}{tools.py (Complete Verbatim)}{Claude Opus 5.5 | \method{} | 554 words | 10 helpers}
import openpyxl
import re
from copy import copy
from openpyxl.worksheet.formula import ArrayFormula


def dump(path='input.xlsx', per_sheet=1500):
    """Print non-empty cells of every sheet (capped). Formulas are shown with their cached values."""
    wb = openpyxl.load_workbook(path)
    wv = openpyxl.load_workbook(path, data_only=True)
    for ws in wb:
        print('==', ws.title, ws.dimensions, 'merged:', [str(m) for m in list(ws.merged_cells.ranges)[:10]])
        n = 0
        for row in ws.iter_rows():
            for c in row:
                v = c.value
                if v is None:
                    continue
                if isinstance(v, ArrayFormula):
                    v = '{' + str(v.text) + '}'
                if isinstance(v, str) and (v.startswith('=') or v.startswith('{=')):
                    print(c.coordinate, repr(v), '->', repr(wv[ws.title][c.coordinate].value))
                else:
                    print(c.coordinate, repr(v))
                n += 1
                if n >= per_sheet:
                    break
            if n >= per_sheet:
                print('...truncated', ws.title)
                break


def last_row(ws, col=None):
    """Real last non-empty row (optionally only in 1-based column col)."""
    for r in range(ws.max_row, 0, -1):
        cells = [ws.cell(r, col)] if col else ws[r]
        if any(c.value is not None and not (isinstance(c.value, str) and not c.value.strip()) for c in cells):
            return r
    return 0


def norm_key(v):
    """Trim, collapse whitespace/NBSP, casefold, and turn 12.0 into '12'."""
    if v is None:
        return None
    if isinstance(v, float) and v.is_integer():
        v = int(v)
    s = re.sub(r'\s+', ' ', str(v).replace('\xa0', ' ')).strip().casefold()
    try:
        f = float(s)
        s = str(int(f)) if f.is_integer() else str(f)
    except ValueError:
        pass
    return s


def find_col(ws, header, row=1):
    """1-based column whose header matches (normalized), else None."""
    h = norm_key(header)
    for c in range(1, ws.max_column + 1):
        if norm_key(ws.cell(row, c).value) == h:
            return c
    return None


def build_lookup(ws, key_cols, val_col, start_row=2, end_row=None):
    """Map a tuple of normalized key-column values to the val_col value. The first occurrence wins."""
    end_row = end_row or last_row(ws)
    d = {}
    for r in range(start_row, end_row + 1):
        k = tuple(norm_key(ws.cell(r, c).value) for c in key_cols)
        d.setdefault(k, ws.cell(r, val_col).value)
    return d


def check_examples(ws, computed, tol=1e-6):
    """Compare computed {coord: value} against existing non-empty cells. Print mismatches; return their count."""
    bad = 0
    for coord, v in computed.items():
        cur = ws[coord].value
        if cur is None or (isinstance(cur, str) and cur.startswith('=')):
            continue
        ok = (abs(float(cur) - float(v)) <= tol) if isinstance(cur, (int, float)) and isinstance(v, (int, float)) else norm_key(cur) == norm_key(v)
        if not ok:
            bad += 1
            if bad <= 10:
                print('MISMATCH', coord, 'have', repr(cur), 'got', repr(v))
    print('example mismatches:', bad)
    return bad


def clear_range(ws, min_row, max_row, min_col, max_col):
    """Set every cell in the range to None."""
    for r in range(min_row, max_row + 1):
        for c in range(min_col, max_col + 1):
            ws.cell(r, c).value = None


def set_font(ws, name=None, **kw):
    """Apply a font name (and other attributes such as bold or size) to every cell in the used range."""
    for row in ws.iter_rows(min_row=1, max_row=ws.max_row, max_col=ws.max_column):
        for c in row:
            f = copy(c.font)
            if name:
                f.name = name
            for k, v in kw.items():
                setattr(f, k, v)
            c.font = f


def insert_col(ws, idx, header=None, header_row=1, fmt_from=None):
    """Insert an empty column at 1-based idx (data shifts right) and optionally set its header.
    fmt_from: source column index, read after the shift, whose header style and number formats are copied."""
    ws.insert_cols(idx)
    if header is not None:
        ws.cell(header_row, idx).value = header
    if fmt_from:
        for r in range(1, ws.max_row + 1):
            s, d = ws.cell(r, fmt_from), ws.cell(r, idx)
            d.number_format = s.number_format
            if r == header_row and s.has_style:
                d.font = copy(s.font)
                d.fill = copy(s.fill)
                d.alignment = copy(s.alignment)


def verify(path, sheet, min_row, max_row, min_col, max_col):
    """Reload the saved file and print the range compactly."""
    ws = openpyxl.load_workbook(path)[sheet]
    for row in ws.iter_rows(min_row=min_row, max_row=max_row, min_col=min_col, max_col=max_col, values_only=True):
        print(row)
\end{pylisting}

\subsection{Gemini 3.8 Flash \method{} Package on \textsc{ALFWorld}}
\label{sec:app_skill_alfworld}

On \textsc{ALFWorld}, agents issue short text commands rather than Python code, so \method{} keeps \texttt{tools.py} empty ($0$ words) and compresses \texttt{SKILL.md} to $656$ words covering search order and object aliases, cutting output tokens by $79.7\%$ ($5{,}704 \to 1{,}155$) at $93.6\%$ test accuracy.

\begin{skilllisting}{SKILL.md (Complete Verbatim)}{Gemini 3.8 Flash | ALFWorld \method{} | 656 words | tools.py: 0 words}
---
name: alfworld_gemini_3.8_pareto
description: Minimal-turn policy for text household tasks (find object, optionally heat/cool/clean, place on target receptacle).
---

# ALFWorld Minimal-Turn Policy

Output exactly one valid action per turn.
- Keep thought to one line at most. Do not deliberate.
- Every pattern below is already solved, so apply it mechanically.

## Valid actions (exact syntax)
- `go to X`, `open X`, `close X`
- `take O from X`
- `put O in/on X` (always the literal "in/on")
- `heat O with microwave N`, `cool O with fridge N`, `clean O with sinkbasin N`

Nothing else works: no slice, use, toggle, examine or look.

## Target object
Only the movable object that ends up on the final receptacle matters.
- **Containers:** in phrases like "pan with Y", "bowl of Y" or "O in the pan", move the contents (Y/O), not the vessel. Fetch the vessel only if it is the only object named.
- **Slices:** slice/piece/cut X means the whole X.
- **Ignore:** adjectives, colors and source phrases.
- **Aliases:** take the first match on sight, including aliases below. If there is no exact name, take the closest synonym on first sight. Never walk past it.

| Instruction word | Object name |
|---|---|
| glass | cup |
| keys | keychain |
| magazine/novel | book |
| tissues | tissuebox |
| sponge | dishsponge |
| soap/bar | soapbar |
| green vegetable | lettuce |
| cell phone | cellphone |
| remote | remotecontrol |
| spray bottle | spraybottle |
| CD | cd |
| TV | television |
| large spoon | ladle/spoon |
| computer | laptop |

## Workflow (4-6 turns)
1. **Find.** Turn 1 is `go to countertop 1`.
   - If the object is not there, visit each location once: fridge (open), diningtable, sidetable/desk, cabinet/drawer (open), microwave, sinkbasin, then other receptacles.
   - Use only receptacle names you have observed.
2. **State change** right after `take`, in place, with no travel:
   - hot/heated/warm/cooked: `heat O with microwave 1`
   - cold/chilled/cooled: `cool O with fridge 1`
   - clean/wash/rinse: `clean O with sinkbasin 1`
3. **Place.** `go to` the destination (open it if closed), then `put` immediately. Stop on "Goal completed successfully!"

## Destination (head noun)
Use the literal receptacle if it exists (cabinet, fridge, sofa, toilet, shelf, etc.). Otherwise map it:

| Instruction word | Receptacle(s), in order |
|---|---|
| table | diningtable, sidetable, desk, coffeetable |
| nightstand/end table | sidetable, diningtable |
| coffee table | coffeetable |
| chair/couch | armchair/sofa |
| bin/trash | garbagecan |
| rack | cart |
| tub | bathtubbasin |
| toilet paper holder | toiletpaperhanger |
| sink | sinkbasin |
| counter | countertop |
| cupboard | cabinet |
| refrigerator | fridge |
| coffee maker | coffeemachine |
| stove | stoveburner |

If nothing matches, use countertop, then diningtable.

## Special patterns
- **Lamp/light tasks:** take X, then put it on diningtable. Never search for a lamp.
- **Two X:** put the first one. Fetch a second only if the goal did not complete.

## Failure handling (strict, no loops)
- **"You can't see any X":** X does not exist. Never retry it or guess other numbers.
- **A put did not complete the goal:** take the object back and go down this ladder once each:
  1. The next mapped destination (another instance or type).
  2. A different state change in place (the required one again if skipped; otherwise cool for food, clean otherwise), then put.
  3. If the object was originally taken from the destination itself, look for another matching object elsewhere, apply the state change, then put.
  4. Stop.
- **Never repeat** a put at the same receptacle more than twice.
- **Never cycle** close/open, re-heat, or revisit an already-checked location.
- **Budget:** about 12 turns. After that, or once every location has been checked, stop.
\end{skilllisting}

\section{Side-by-Side Trajectory Comparisons Across Benchmarks}
\label{sec:appendix_forensics}

We compare held-out test trajectories under \evoskill{} ($A^*$) and \method{} ($\mathcal{S}^*$) on the same tasks across \textsc{SpreadsheetBench}, \textsc{ALFWorld}, and \textsc{MATH-500} (\S\ref{sec:diagnostics}).

\subsection{\textsc{SpreadsheetBench}: Stopping No-Code Loops \& Unsaved Submissions}
\label{sec:app_trace_wandering}

Across $12{,}130$ evaluated \textsc{SpreadsheetBench} turns ($10{,}730$ in the $4 \times 2 \times 280$ main test matrix plus seed and ablation runs), two failure modes drive most accuracy losses under verbose, helper-free skills. First, on Gemini 3.8 Flash, \evoskill{}'s $1{,}271$-word \texttt{SKILL.md} and empty \texttt{tools.py} leave $26.1\%$ of turns ($763/2{,}923$) without executable code ($27\%$ following another no-code turn) while inline scripting errors raise next-turn error rates from $5.8\%$ to $10.4\%$; \method{} cuts unexecuted turns to $14.1\%$ ($338/2{,}398$: $311$ code-free completion turns and $27$, or $1.1\%$, non-empty prose turns) and cuts $20$-turn timeouts by $60\%$ ($45 \to 18$). Second, on Grok 4.1 Fast, long Turn-1 plans under \evoskill{} emit the completion signal before saving on $51/280$ runs ($18.2\%$, $0\%$ pass); \method{} cuts unsaved exits to $9/280$ ($3.2\%$) and recovers $40$ of those $51$ tasks ($+49$-task net gain).

\paragraph{No-Code Loop vs.\ Clean 6-Turn Solve (Gemini 3.8 Flash, Task \texttt{15380}).}
Task \texttt{15380} extracts two-digit campaign codes from \texttt{A3:A14} (e.g., \texttt{"2014\_09\_D2\_Reminder\_EN"} $\to$ \texttt{"09"}), maps them via named ranges \texttt{codes2} (\texttt{D3:D14}) and \texttt{month} (\texttt{E3:E14}), and writes month labels (\texttt{"SEP"}, \dots, \texttt{"NOV"}) to \texttt{B3:B14}; a naive \texttt{code in text} check tests \texttt{"01"} first and falsely matches \texttt{"2014"} (\texttt{"JAN"}). In Table~\ref{tab:paired_trajectory_15380}, Gemini 3.8 Flash under \evoskill{} hits this bug at Turn~6, stalls across $9$ prose-only turns ($22{,}852$ output tokens), and times out at Turn~20 without saving. Under \method{}, mandatory-code rules keep every turn executable (even when Turn~1 queries \texttt{openpyxl} directly rather than \texttt{inspect\_wb}), catching the \texttt{"2014"} collision at Turn~3 and saving verified labels by Turn~6 ($-68.1\%$ output, $-81.0\%$ input tokens). On the same task, Claude Opus 5.5 with \method{} (Appendix~\ref{sec:app_skill_opus_sheet}) finishes in $3$ turns ($468$ output, $7{,}954$ input tokens: $16$-tok inspect, $440$-tok compute-and-verify, $12$-tok submit).

\begin{table}[!htbp]
\centering
\footnotesize
\setlength{\tabcolsep}{5pt}
\renewcommand{\arraystretch}{1.03}
\caption{Paired trajectory on \textsc{SpreadsheetBench} Task \texttt{15380} for Gemini 3.8 Flash. ``Out Tok'' reports total output tokens (reasoning $+$ visible completion) per turn or turn range.}
\label{tab:paired_trajectory_15380}
\begin{tabularx}{\textwidth}{@{}l >{\raggedright\arraybackslash}X r >{\raggedright\arraybackslash}X r@{}}
\toprule
& \multicolumn{2}{l}{\textbf{\evoskill{} (Unconstrained, $A^*$)}} & \multicolumn{2}{l}{\textbf{\method{} (Ours, $\mathcal{S}^*$)}} \\
& \multicolumn{2}{l}{\scriptsize\color{warnCrimson}\textbf{FAIL}: 20 turns, 36,669 out tok, 134,553 in tok} & \multicolumn{2}{l}{\scriptsize\color{gainForest}\textbf{PASS}: 6 turns, 11,708 out tok, 25,629 in tok} \\
\cmidrule(lr){2-3} \cmidrule(lr){4-5}
\textbf{Turn} & \textbf{Action \& Workspace Outcome} & \textbf{Out Tok} & \textbf{Action \& Workspace Outcome} & \textbf{Out Tok} \\
\midrule
\textbf{T1} & Prints sheet names and rows 2--5 of \texttt{Sheet3}. & 595 & Loads workbook; queries \texttt{defined\_names.\hspace{0pt}definedName} $\to$ \texttt{AttributeError}. & 327 \\
\textbf{T2} & Re-prints \texttt{A1:G20} coordinate tuples. & 355 & Fixes query; resolves \texttt{codes2} (\texttt{D3:D14}) and \texttt{month} (\texttt{E3:E14}). & 615 \\
\textbf{T3} & Inspects font, fill, and borders of empty col \texttt{B}. & 1,240 & Builds \texttt{'01'..'12'} map; sees naive match hit \texttt{'01'} in \texttt{'2014'} (\texttt{JAN}). & 7,480 \\
\textbf{T4} & \textcolor{warnCrimson}{\emph{Prose only (no code executed).}} & 3,565 & Verifies \texttt{str} types across \texttt{D2:E14} before writing. & 932 \\
\textbf{T5} & \textcolor{warnCrimson}{\emph{Prose only (no code executed).}} & 5,337 & Extracts exact code, writes \texttt{'SEP'..'NOV'} to \texttt{B3:B14}, saves workbook. & 1,995 \\
\textbf{T6} & Tests naive match (\texttt{JAN}); prints dict without saving. & 1,201 & Issues completion signal $\to$ \textcolor{gainForest}{\textbf{Pass (verified)}}. & 359 \\
\textbf{T7--T8} & Re-checks \texttt{type(D3:D14)}; \textbf{T8} \textcolor{warnCrimson}{\emph{prose only}}. & 2,074 & \multicolumn{2}{c}{\emph{\color{inkMuted}(Episode finished at Turn 6)}} \\
\textbf{T9--T11} & Re-dumps \texttt{MID} slices, cell styles, and named ranges. & 4,137 & & \\
\textbf{T12--T15} & \textcolor{warnCrimson}{\emph{Four consecutive prose-only turns}} (no code run). & 6,643 & & \\
\textbf{T16--T17} & Re-prints \texttt{A2:B10}; \textbf{T17} \textcolor{warnCrimson}{\emph{prose only}}. & 3,629 & & \\
\textbf{T18--T19} & \textbf{T18} \texttt{AttributeError} on \texttt{definedName}; \textbf{T19} \textcolor{warnCrimson}{\emph{prose only}}. & 4,911 & & \\
\textbf{T20} & Script prints nothing, never saves $\to$ \textcolor{warnCrimson}{\textbf{Timeout}}. & 2,982 & & \\
\midrule
\textbf{Total} & \scriptsize 32,027 reasoning + 4,642 visible (9 unexecuted turns) & \textbf{36,669} & \scriptsize 10,070 reasoning + 1,638 visible (0 unexecuted turns) & \textbf{11,708} \\
\bottomrule
\end{tabularx}
\end{table}

\paragraph{Unsaved Turn-1 Exit vs.\ Verified Save (Grok 4.1 Fast, Task \texttt{36191}).}
Task \texttt{36191} strips trailing emails from names in \texttt{Column B} (e.g., \texttt{"Sam Landon sam@BTR.com"} $\to$ \texttt{"Sam Landon"}) and writes them to \texttt{Column C}. Under \evoskill{}, Grok 4.1 Fast generates $14{,}649$ Turn-1 output tokens ($14{,}152$ visible): its inline inspection script crashes, yet its trailing plan emits the completion signal and ends on an untouched file (Table~\ref{tab:paired_trajectory_grok_36191}). Under \method{}, the \texttt{tools.py} inspection helper prints \texttt{A1:C3} and reminds the agent that no edits are saved yet, yielding a verified $2$-turn solve ($-82.3\%$ output tokens).

\begin{table}[!htbp]
\centering
\footnotesize
\setlength{\tabcolsep}{5pt}
\renewcommand{\arraystretch}{1.03}
\caption{Paired trajectory on \textsc{SpreadsheetBench} Task \texttt{36191} for Grok 4.1 Fast.}
\label{tab:paired_trajectory_grok_36191}
\begin{tabularx}{\textwidth}{@{}l >{\raggedright\arraybackslash}X r >{\raggedright\arraybackslash}X r@{}}
\toprule
& \multicolumn{2}{l}{\textbf{\evoskill{} (Unconstrained, $A^*$)}} & \multicolumn{2}{l}{\textbf{\method{} (Ours, $\mathcal{S}^*$)}} \\
& \multicolumn{2}{l}{\scriptsize\color{warnCrimson}\textbf{FAIL}: 1 turn, 14,649 out tok, 2,173 in tok} & \multicolumn{2}{l}{\scriptsize\color{gainForest}\textbf{PASS}: 2 turns, 2,597 out tok, 3,940 in tok} \\
\cmidrule(lr){2-3} \cmidrule(lr){4-5}
\textbf{Turn} & \textbf{Action \& Workspace Outcome} & \textbf{Out Tok} & \textbf{Action \& Workspace Outcome} & \textbf{Out Tok} \\
\midrule
\textbf{T1} & Inlines custom inspection script (\texttt{Traceback}) and narrates full plan ending in completion signal $\to$ \textcolor{warnCrimson}{\textbf{Early exit (0 saves)}}. & 14,649 & Calls \texttt{tools.py} inspection helper; prints \texttt{A1:C3} and reminder that no edits are saved yet. & 226 \\
\textbf{T2} & \multicolumn{2}{c}{\emph{\color{inkMuted}(Ended without saving at Turn 1)}} & Strips email suffix, writes \texttt{"Sharon Smith"} and \texttt{"Sam Landon"} to \texttt{C2:C3}, saves, verifies, completes $\to$ \textcolor{gainForest}{\textbf{Pass}}. & 2,371 \\
\midrule
\textbf{Total} & \scriptsize 497 reasoning + 14,152 visible completion & \textbf{14,649} & \scriptsize 2,429 reasoning + 168 visible completion & \textbf{2,597} \\
\bottomrule
\end{tabularx}
\end{table}

\subsection{\textsc{ALFWorld}: Targeted Object Search vs.\ Wandering Across Rooms}
\label{sec:app_alfworld_trajectory}

On \textsc{ALFWorld}, actions are short ($4$--$15$ tokens), so token cost comes mainly from per-turn reasoning and extra room visits. In Table~\ref{tab:paired_trajectory_alfworld} (\texttt{T20190911\_131350\_027076}, \emph{``put clean spoon on the table''}), Gemini 3.8 Flash under \evoskill{} sees \texttt{ladle 1} on \texttt{countertop 1} at Turn~2 but walks past it looking for a literal \texttt{spoon}, checks $7$ other receptacles (Turns~3--11) and $4$ nonexistent IDs (Turns~12--15), and spends $5{,}083$ reasoning tokens before returning at Turn~16. Under \method{} (Appendix~\ref{sec:app_skill_alfworld}), three simple rules (start at \texttt{countertop 1}, match aliases like \texttt{spoon} $\to$ \texttt{ladle} on sight, and run \texttt{clean} in place) finish in $5$ turns, cutting output tokens by $96.3\%$ ($5{,}206 \to 195$) and input tokens by $86.8\%$ ($49{,}167 \to 6{,}479$).

Ordered fallback locations also lift Claude Opus 5.5 accuracy from $88.9\%$ to $93.2\%$ ($12$ failed test tasks recovered, $0$ lost). On \texttt{T20190908\_210847\_591144} (\emph{``Place a warm cup on the counter where the microwave is''}, with the microwave on \texttt{sidetable 1}), \evoskill{} repeatedly re-heats \texttt{cup 1} and \texttt{mug 1} until timing out at Turn~25 ($3{,}448$ output, $87{,}486$ input tokens), whereas \method{} heats \texttt{cup 1} once and checks \texttt{countertop 1} $\to$ \texttt{diningtable 1} $\to$ \texttt{sidetable 1} in order to pass in $11$ turns ($590$ output, $17{,}236$ input tokens).

\begin{table}[!htbp]
\centering
\footnotesize
\setlength{\tabcolsep}{5pt}
\renewcommand{\arraystretch}{1.03}
\caption{Paired trajectory on \textsc{ALFWorld} task \texttt{T20190911\_131350\_027076} (\emph{``put clean spoon on the table''}) for Gemini 3.8 Flash.}
\label{tab:paired_trajectory_alfworld}
\begin{tabularx}{\textwidth}{@{}l >{\raggedright\arraybackslash}X r >{\raggedright\arraybackslash}X r@{}}
\toprule
& \multicolumn{2}{l}{\textbf{\evoskill{} (Unconstrained, $A^*$)}} & \multicolumn{2}{l}{\textbf{\method{} (Ours, $\mathcal{S}^*$)}} \\
& \multicolumn{2}{l}{\scriptsize\color{gainForest}\textbf{PASS}: 21 turns, 5,206 out tok, 49,167 in tok} & \multicolumn{2}{l}{\scriptsize\color{gainForest}\textbf{PASS}: 5 turns, 195 out tok, 6,479 in tok} \\
\cmidrule(lr){2-3} \cmidrule(lr){4-5}
\textbf{Turn} & \textbf{Action \& Environment Observation} & \textbf{Out Tok} & \textbf{Action \& Environment Observation} & \textbf{Out Tok} \\
\midrule
\textbf{T1} & \texttt{go to diningtable 1} $\to$ empty. & 115 & \texttt{go to countertop 1} $\to$ sees \texttt{knife 1, mug 1, ladle 1}. & 24 \\
\textbf{T2} & \texttt{go to countertop 1} $\to$ sees \texttt{ladle 1}, \textcolor{warnCrimson}{\emph{walks past it}}. & 75 & \texttt{take ladle 1 from countertop 1} (\texttt{spoon} $\to$ \texttt{ladle} alias). & 44 \\
\textbf{T3} & \texttt{go to sinkbasin 1} $\to$ empty. & 138 & \texttt{clean ladle 1 with sinkbasin 1} (in-place state change). & 41 \\
\textbf{T4--T5} & Visits and opens \texttt{cabinet 1} $\to$ sees \texttt{plate 1}. & 159 & \textbf{T4:} \texttt{go to diningtable 1} (\texttt{table} $\to$ \texttt{diningtable}). \newline \textbf{T5:} \texttt{put ladle 1 in/on diningtable 1} $\to$ \textcolor{gainForest}{\textbf{Goal completed!}} & 86 \\
\textbf{T6--T11} & Opens/checks \texttt{fridge 1}, \texttt{microwave 1}, \texttt{coffeemachine 1}, and \texttt{stove 1} without finding a literal \texttt{spoon}. & 1,475 & \multicolumn{2}{c}{\emph{\color{inkMuted}(Episode finished at Turn 5; T4: 61 tok, T5: 25 tok)}} \\
\textbf{T12--T15} & Guesses non-existent \texttt{drawer 1}, \texttt{cabinet 2}, \texttt{garbagecan 1}, \texttt{countertop 2} (\emph{``can't see any \dots''}). & 1,495 & & \\
\textbf{T16--T17} & Returns to \texttt{countertop 1} and takes \texttt{ladle 1} after long reasoning. & 1,284 & & \\
\textbf{T18--T19} & Travels to \texttt{sinkbasin 1}; cleans \texttt{ladle 1}. & 245 & & \\
\textbf{T20--T21} & Travels to \texttt{diningtable 1}; places \texttt{ladle 1} $\to$ \textcolor{gainForest}{\textbf{Goal completed!}} & 220 & & \\
\midrule
\textbf{Total} & \scriptsize 5,083 reasoning + 123 action tokens & \textbf{5,206} & \scriptsize 156 reasoning + 39 action tokens & \textbf{195} \\
\bottomrule
\end{tabularx}
\end{table}

\subsection{\textsc{MATH-500}: Direct Symbolic Solution vs.\ Repeated Code Retries}
\label{sec:app_math500_derivation}

Across $10{,}304$ evaluated \textsc{MATH-500} runs ($2{,}240$ in the $4 \times 2 \times 280$ test matrix plus validation and ablation runs), single-turn symbolic solutions pass $78.7\%$ at $3{,}363$ total tokens per solve, whereas multi-turn Python/SymPy runs pass $75.4\%$ at $15{,}316$ tokens ($4.6\times$ higher, reflecting both harder problems when models call Python and retry overhead). \method{} shifts Gemini 3.8 Flash ($\bar{T}: 1.76 \to 1.04$) and Gemma 4 26B ($\bar{T}: 1.70 \to 1.19$) from code retries to direct Turn-1 derivations while trimming prose restatements.

On \texttt{test\_precalculus\_190} ($\prod_{k=1}^{8} \sin(20^\circ k) = \frac{9}{256}$; Table~\ref{tab:paired_trajectory_math500}), Gemini 3.8 Flash under \evoskill{} spends $4$ turns across unreduced SymPy output, a no-code pause, a float check, and a long write-up ($2{,}496$ output, $9{,}946$ input tokens); under \method{}, it applies trigonometric identities directly in $1$ turn ($471$ output, $1{,}775$ input tokens; $-81.1\%$ and $-82.2\%$). This also avoids sandbox syntax-error loops on open-weight models: on \texttt{test\_intermediate\_algebra\_1354} ($\sum_{k=1}^{1997}\frac{1}{1+\omega^k}$ for $\omega^{1997}=1, \omega\neq 1$), Gemma 4 26B fails across $10$ malformed tool calls under \evoskill{} ($2{,}336$ output, $45{,}580$ input tokens) versus a $1$-turn symbolic solve under \method{} ($544$ output, $1{,}643$ input tokens). On single-turn Claude Opus 5.5, \method{} strips extra prose ($792 \to 100$ output tokens on \texttt{test\_precalculus\_190}; $522 \to 33$ on \texttt{test\_intermediate\_algebra\_1247}).

\begin{table}[!htbp]
\centering
\footnotesize
\setlength{\tabcolsep}{5pt}
\renewcommand{\arraystretch}{1.03}
\caption{Paired trajectory on \textsc{MATH-500} problem \texttt{test\_precalculus\_190} ($\sin 20^\circ \sin 40^\circ \dots \sin 160^\circ$, Gemini 3.8 Flash).}
\label{tab:paired_trajectory_math500}
\begin{tabularx}{\textwidth}{@{}l >{\raggedright\arraybackslash}X r >{\raggedright\arraybackslash}X r@{}}
\toprule
& \multicolumn{2}{l}{\textbf{\evoskill{} (Unconstrained, $A^*$)}} & \multicolumn{2}{l}{\textbf{\method{} (Ours, $\mathcal{S}^*$)}} \\
& \multicolumn{2}{l}{\scriptsize\color{gainForest}\textbf{PASS}: 4 turns, 2,496 out tok, 9,946 in tok} & \multicolumn{2}{l}{\scriptsize\color{gainForest}\textbf{PASS}: 1 turn, 471 out tok, 1,775 in tok} \\
\cmidrule(lr){2-3} \cmidrule(lr){4-5}
\textbf{Turn} & \textbf{Action \& Sandbox Outcome} & \textbf{Out Tok} & \textbf{Action \& Symbolic Solution} & \textbf{Out Tok} \\
\midrule
\textbf{T1} & Runs SymPy \texttt{sp.simplify(sp.prod(...))} $\to$ returns unreduced radian product. & 570 & Pairs $\sin(180^\circ-\theta)=\sin\theta$ into $(\sin 20^\circ\sin 40^\circ\sin 60^\circ\sin 80^\circ)^2$, applies $\frac{1}{4}\sin(3\theta)$ identity $\to$ $(\frac{3}{16})^2 = \frac{9}{256}$ $\to$ \textcolor{gainForest}{\textbf{Pass}}. & 471 \\
\textbf{T2} & Emits prose without code block $\to$ no output. & 166 & \multicolumn{2}{c}{\emph{\color{inkMuted}(Episode finished at Turn 1)}} \\
\textbf{T3} & Runs Python float product (\texttt{0.03515625}) and checks \texttt{Fraction(3,16)**2 = 9/256}. & 696 & & \\
\textbf{T4} & Writes full multi-paragraph prose proof and final answer $\frac{9}{256}$ $\to$ \textcolor{gainForest}{\textbf{Pass}}. & 1,064 & & \\
\midrule
\textbf{Total} & \scriptsize 1,527 reasoning + 969 visible completion & \textbf{2,496} & \scriptsize 325 reasoning + 146 visible completion & \textbf{471} \\
\bottomrule
\end{tabularx}
\end{table}

\end{document}